\documentclass[APA,Times2COL]{WileyNJDv5} 

\articletype{Article Type}%

\received{Date Month Year}
\revised{Date Month Year}
\accepted{Date Month Year}
\journal{Journal}
\volume{00}
\copyyear{2023}
\startpage{1}

\begin{document}

\title{Real-time Sampling-based Model Predictive Control based on Reverse Kullback-Leibler Divergence and Its Adaptive Acceleration}

\author[1,2]{Taisuke Kobayashi}

\author[3]{Kota Fukumoto}

\authormark{KOBAYASHI \textsc{et al.}}
\titlemark{Real-time Sampling-based Model Predictive Control based on Reverse Kullback-Leibler Divergence and Its Adaptive Acceleration}

\address[1]{\orgdiv{Principles of Informatics Research Division}, \orgname{National Institute of Informatics (NII)}, \orgaddress{\street{2-1-2 Hitotsubashi}, \city{Chiyoda-ku}, \postcode{101-8430}, \state{Tokyo}, \country{Japan}}}

\address[2]{\orgdiv{Informatics Program}, \orgname{The Graduate University for Advanced Studies (SOKENDAI)}, \orgaddress{\street{2-1-2 Hitotsubashi}, \city{Chiyoda-ku}, \postcode{101-8430}, \state{Tokyo}, \country{Japan}}}

\address[3]{\orgdiv{Division of Information Science}, \orgname{Nara Institute of Science and Technology (NAIST)}, \orgaddress{\street{8916-5 Takayama-cho}, \city{Ikoma}, \postcode{630-0192}, \state{Nara}, \country{Japan}}}

\corres{Corresponding author: Taisuke Kobayashi \email{kobayashi@nii.ac.jp}}

\abstract[Abstract]{%
Sampling-based model predictive control (MPC) has the potential for use in a wide variety of robotic systems.
However, its unstable updates and poor convergence render it unsuitable for real-time control of robotic systems.
This study addresses this challenge with a novel approach from reverse Kullback-Leibler divergence, which has a mode-seeking property and is likely to find one of the locally optimal solutions early.
Using this approach, a weighted maximum likelihood estimation with positive and negative weights is obtained and solved using the mirror descent (MD) algorithm.
Negative weights eliminate unnecessary actions, but a practical implementation needs to be designed to avoid interference with positive and negative updates based on rejection sampling.
In addition, Nesterov's acceleration method for the proposed MD is modified to improve heuristic step size adaptive to the noise estimated in update amounts.
Real-time simulations show that the proposed method can solve a wider variety of tasks statistically than the conventional method.
In addition, higher degrees-of-freedom tasks can be solved by the improved acceleration even with a CPU only.
The real-world applicability of the proposed method is also demonstrated by optimizing the operability in a variable impedance control of a force-driven mobile robot.
}


\maketitle

\section{Introduction}

Increasingly complex robotic systems with various, such as disaster response~\cite{delmerico2019current,kobayashi2015selection} and physical human-robot interaction~\cite{ajoudani2018progress,kobayashi2021whole}, require more versatile controllers.
Data-driven control techniques, such as imitation learning~\cite{osa2018algorithmic} and reinforcement learning~\cite{sutton2018reinforcement}, have gained attention.
However, the former requires skilled data for each task, and the latter involves a large amount of trial and error, making it difficult to apply either method to any task.

In contrast, MPC~\cite{camacho2013model} can accomplish the desired tasks by minimizing or maximizing the task-specific cost (or reward) function provided that the model of the target system is known.
MPC has been successfully implemented in various robot control problems such as cutting foods~\cite{lenz2015deepmpc}, autonomous driving~\cite{cui2021autonomous}, walking on rough terrain~\cite{gibson2021terrain}, and etc.
This study focuses on sampling-based MPC~\cite{botev2013cross,williams2018information}, which selects the best control input/action from a policy distribution.

Sampling-based MPC is more applicable to systems with non-differentiable (even discontinuous) dynamics and cost functions.
However, it is known for its poor convergence and high computational demands, lacking of real-time performance~\cite{pinneri2021sample}.
This is a significant obstacle for robot control, in which real-time performance is often required.
Thereby, various measures have been considered to address this issue (details are in the next section).

Here, we focus on the following two previous studies, which are closely related to this paper.
One study developed a method that evaluates more important action candidates for solving the given task by utilizing the concept of importance sampling~\cite{bishop2006pattern}, while restricting the exploration area of the action space (a.k.a. the scale of policy)~\cite{carius2022constrained}.
As this method efficiently samples superior action candidates, it can improve convergence speed.
However, it was heuristically designed with a parameterized policy that was trained numerically without analytical derivation.
Another study incorporated Nesterov's accelerated gradient (NAG)~\cite{nesterov1983method} into MPC to speed up convergence~\cite{okada2018acceleration}.
The introduction of this momentum-based acceleration method has been well established in recent deep-learning optimizers~\cite{zhang2019lookahead,ilboudo2023adaterm} and its effectiveness is widely known.
However, this method updates the location of policy and does not optimize its scale.

Inspired by these two ideas ---the selective action sampling and the momentum-based acceleration---, this study aims to develop a novel sampling-based MPC that is theoretically grounded and implemented to perform as per its theory.
The optimization problem of conventional sampling-based MPC~\cite{botev2013cross,williams2018information} is revised to facilitate the preferential selection of superior action candidates while avoiding inferior ones.
Specifically, a new optimization problem starts by minimizing \textit{reverse} Kullback-Leibler (RKL) divergence between the estimated optimal policy and the policy to be updated, instead of \textit{forward} Kullback-Leibler (FKL) divergence conventionally adopted.
This problem can be solved by mirror descent (MD) algorithm~\cite{beck2003mirror}, in which the policy's location and scale are theoretically optimized such that the more/less valuable candidates are actively selected/avoided.

The update rule employs two vectors: one for approaching superior candidates, and the other for leaving inferior ones.
However these vectors can interfere with each other; therefore, a practical implementation is required to mitigate this interference.
This implementation decomposes the policy into two parts for generating important and worthless candidates.
Both policies are optimized separately and then combined using rejection sampling~\cite{bishop2006pattern}.
This ensures that only superior candidates are accepted, whereas worthless ones are rejected.

Moreover, the proposed MPC is integrated with NAG.
The latest NAG for MD~\cite{cohen2018acceleration} is modified to fit the proposed method.
In addition, motivated by the report that NAG is sensitive to noise in update amounts~\cite{cohen2018acceleration}, a heuristic is developed to estimate the noise from the respective candidates' outcomes and slow down the acceleration accordingly.
When applied to optimize the exploration space, this adaptive acceleration makes exploration and convergence more efficient simultaneously.

The proposed method is validated step by step.
First, the performance of the proposed method is statistically verified without adaptive acceleration by simulating simple tasks on the Brax simulator~\cite{freeman2021brax} on a GPU.
The results showed that the proposed method can solve more tasks than the conventional method.
Second, to verify the effectiveness of the adaptive acceleration method, more complex tasks are solved using only a CPU on the Brax simulator.
Despite conservative updates to alleviate the instability associated with fewer candidates limited by computer resources, only the proposed acceleration can successfully solve the tasks.
Finally, the entire proposed method is applied to a variable impedance control of a force-driven mobile robot~\cite{kobayashi2022light} to demonstrate its applicability to real-time robotic systems (in this case, 50~Hz is required only with an embedded CPU).

This paper's contributions are two theoretical (T1 and T2) and two heuristic (H1 and H2) aspects as follows:
\begin{description}
    \item[T1:] Derivation of a new MPC, so-called Reverse MPC, starting from minimization of RKL divergence.
    \item[H1:] Design of a heuristic called Reject MPC, which mitigates the interference of Reverse MPC.
    \item[T2:] Appropriate modification to apply the latest NAG on MD into Reject MPC called Accel MPC.
    \item[H2:] Design of adaptive acceleration/slowdown to mitigate the noise sensitivity of Accel MPC.
\end{description}
Simulations have statistically validated the effectiveness of T1 and H1.
Moreover, simulations and demonstrations of T2 and H2 demonstrated their practicality for real-time robot control, even with limited computational resources.

\subsection{Related work}

\subsubsection{Types of nonlinear MPC}

MPC was initially limited to obtaining feedback gains only for linear systems, but later developments have focused on numerical solutions for nonlinear systems.
There are two main streams of nonlinear MPC:
A gradient-based approach takes advantage of the gradient of model and cost function~\cite{tassa2014control,ohtsuka2004continuation};
and a sampling-based approach that evaluates many action candidates to find the optimal one~\cite{botev2013cross,williams2018information}.
Note that these methods have been designed for practical use, and even a well-converged numerical solution for a given optimal problem does not guarantee stability.
Therefore, the method proposed in this study also pursue the same direction without discussing stability in depth.

\paragraph{Gradient-based approach}
The gradient-based approach typically uses iterative linear quadratic regulator (iLQR)~\cite{tassa2014control} and continuation and generalized minimal residual method (C/GMRES)~\cite{ohtsuka2004continuation} solvers, which provide relatively fast convergence.
This approach is often used in robotics, where real-time performance is essential.
However, it does not apply to non-differentiable systems because of the assumption of differentiability in the target.
This study, therefore, ignored this approach to make the applicable system classes as wide as possible.

\paragraph{Sampling-based approach}
The sampling-based approach typically uses the cross-entropy method (CEM)~\cite{botev2013cross} and model predictive path integral (MPPI)~\cite{williams2018information} methods, both of which are versatile because they allow the target model and cost function to be discontinuous.
Therefore, they are widely employed in model-based reinforcement learning~\cite{chua2018deep,deisenroth2011pilco},
where the model (and cost function) is accurately learned from data, such as deep learning~\cite{chua2018deep} or Gaussian processes~\cite{deisenroth2011pilco}, and utilized in MPC.

\subsubsection{Improvement of real-time performance}

Numerous studies have aimed to improve the real-time performance of sampling-based MPC.
In this study, we briefly introduced them by dividing them into several categories.
However, we must remark that they enhance the real-time performance of MPC in practice, but does not improve MPC itself from a theoretical perspective.

\paragraph{Efficient exploration}
One approach is to use a pretrained expressive model to generate action candidates. Although this increases computational cost due to the complexity of the learned model~\cite{kusumoto2019informed,sacks2023learning}, it provides efficient exploration.
While it is not applied in real-robot applications, the literature~\cite{pinneri2021sample} introduced various heuristics to improve the real-time performance of CEM, such as utilizing colored noise for exploration and reusing previously good candidates.
These two approaches can be used complementarily in this study if MD can be solved for them.

\paragraph{Extraction of lower dimension}
Another approach aims to reduce the computational cost by making the model as low-dimensional as possible~\cite{kobayashi2023sparse}.
Although not using sampling-based MPC, humanoid robot control was efficiently achieved by dividing the robot model into two hierarchies according to its responsiveness and optimizing them with iLQR step by step~\cite{ishihara2019full}.
These two approaches can be used as complements to this study.

\paragraph{Use of gradient-based method}
A method that combines sampling-based MPC with a gradient-based approach was proposed~\cite{bharadhwaj2020model} to compensate for the slow convergence of MPC alone.
Another approach is to implement or approximate in a differentiable form to compute gradients for any task~\cite{amos2018differentiable}.
However, the former has limitations regarding its application, and the latter may have more approximation errors when it includes discrete events.

\paragraph{Improvements at implementation level}
In the case study of MPPI applied to Auto-Rally~\cite{williams2018information}, a relatively simple model was described, and parallel computations on a GPU were used to ensure real-time performance.
In the case study of CEM applied to quadruped locomotion~\cite{yang2020data}, a low-level controller with real-time property and a high-level controller with actions from CEM were implemented asynchronously to mitigate the delay of CEM optimization.
As the time delay in MPC optimization is unavoidable in real-time control, an implementation that compensates for it by predicting the initial state of the optimization~\cite{cui2021autonomous} has been proposed.

\section{Problem statement}

\subsection{Sampling-based model predictive control}

In general, MPC targets nonlinear discrete-time deterministic system $f(x, u)$ with state $x \in \mathcal{X} \subset \mathbb{R}^{|\mathcal{X}|}$ and action $u \in \mathcal{U} \subset \mathbb{R}^{|\mathcal{U}|}$, sometimes under constraint $C(x, u)$~\cite{camacho2013model}.
At discrete-time step $t$ with $x_t$, MPC optimizes the action sequence $U_t = \{ u_t, \dots , u_{t+H} \}$ from $t$ to $t+H$ with $H$ the horizon length.
This optimization problem is given as follows:
\begin{align}
  U_t^{\ast} &= \{ u_t^{\ast}, \dots , u_{t+H}^{\ast} \} = \arg \min_{U_t} J(x_t, U_t)
  \label{eq:mpc_prob} \\
  s.t.\ &J(x_t, U_t) = \phi(x_{t+H+1}) + \sum^{t+H}_{\tau=t} L(x_\tau, u_\tau)
  \nonumber \\
  &x_{\tau+1} = f(x_\tau, u_\tau), \ \tau = t,\ldots,t+H
  \nonumber \\
  &C(x_\tau, u_\tau) \le 0, \ \tau = t,\ldots,t+H
  \nonumber
\end{align}
where $J(x, U) \in \mathbb{R}$ is the cost function, given by the sum of the termination cost $\phi(x) \in \mathbb{R}$ and the stage cost $L(x, u) \in \mathbb{R}$.
If $C(x, u)$ cannot be directly handled like the sampling-based MPC, it will be converted into a large cost when violated.
This policy was adopted in this study to relax the constraint.
After this optimization, $u_t^{\ast}$ is extracted from $U_t^{\ast}$ to transition the real system to the next state $x_{t+1}$.
MPC repeats the above optimization at each time step.

\begin{figure}[tb]
    \centering
    \includegraphics[keepaspectratio=true,width=0.96\linewidth]{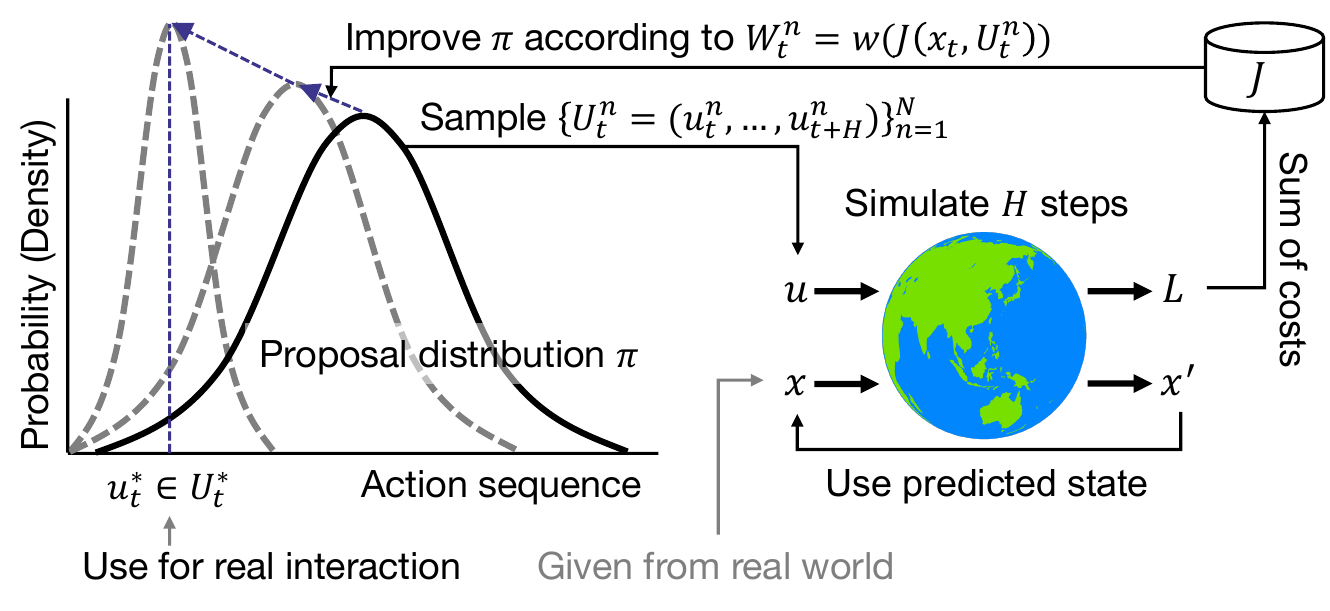}
    \caption{Overview of sampling-based MPC.
    Candidates of action sequence are sampled from $\pi$ and are evaluated using the given model.
    According to the sum of costs of each candidate, $\pi$ is updated to selectively sample the candidates with smaller costs.
    By repeating this process, $\pi$ converges to the optimal solution.
    }
    \label{fig:framework_mpc}
\end{figure}

Several solvers have been proposed for this optimization problem, including PDIP and SQP for general nonlinear optimization solvers~\cite{nocedal1999numerical}; and iLQR~\cite{tassa2014control} and C/GMRES~\cite{ohtsuka2004continuation} for computationally-fast solvers.
Several optimizations limit the functions of $f(x, u)$ and $J(x, U)$ to differentiable classes.
This study focused on the sampling-based MPC approach, as illustrated in Fig.~\ref{fig:framework_mpc}.
Note that for the sake of simplicity, we discuss deterministic systems in this study although this method can be applied to stochastic systems.
In this approach, the proposed distribution (or policy) is introduced $\pi(U; \theta)$ with $\theta$ the model parameters set.
This distribution stochastically generates $U$.

Note that $\pi(U; \theta)$ is often modeled as a diagonal normal distribution, with in~\cite{williams2018information,okada2020variational} (i.e. $\theta = [\mu \in \mathbb{R}^{|\mathcal{U}| \times H}, \sigma \in \mathbb{R}^{|\mathcal{U}| \times H}_{+}]$).
As the sampled element $u \in U$, however, may not satisfy the input constraints, a nonlinear transformation is incorporated into the system, mapping $\mathbb{R}^{|\mathcal{U}|} \to \mathcal{U}$.
Specifically, when input constraints are given as upper and lower bounds, $u \in [\underline{u}, \overline{u}]]$, the following transformation is used in this study.
\begin{align}
    \tilde{u} &= \frac{1}{2} \{\tanh(u) (\overline{u} - \underline{u}) + \overline{u} + \underline{u}\}
\end{align}
This transformed $\tilde{u} \in \mathcal{U}$ can be input to the system (and the cost function).

From $\pi(U_t; \theta)$ at $t$, the sampling-based MPC (represented by CEM~\cite{botev2013cross}) stochastically samples $N$ candidates of the action sequences, $\{U_t^n\}_{n=1}^N$.
$J(x_t, U_t^n)$ can be computed for each candidate as straightforward.
Then, the optimal weight of each candidate $W_t^n$ (i.e. higher is better) is evaluated with a map $w: \mathbb{R} \to \mathbb{R}_{+}$ as follows:
\begin{align}
    W_t^n = w(J(x_t, U_t^n))
    \label{eq:mpc_score}
\end{align}
Note that several solvers can be categorized based on the type of $w$, as indicated in the literature~\cite{okada2020variational}.
For instance, CEM employs a threshold-based $w$~\cite{botev2013cross} while MPPI uses exponential-based $w$~\cite{williams2018information}.
\begin{align}
    &W_t^n \propto
    \label{eq:mpc_weight_fkl} \\
    &\begin{cases}
        \mathbb{I}(J(x_t, U_t^n) \leq \mathrm{Q}(\{J(x_t, U_t^n)\}_{n=1}^N; \lambda)) & \mathrm{CEM}
        \\
        \exp(-J(x_t, U_t^n) / T) & \mathrm{MPPI}
    \end{cases}
    \nonumber
\end{align}
where $\mathbb{I}(\cdot)$ is the indicator function that outputs one if the given condtion holds, otherwise zero.
In addition, $\lambda \in (0, 1)$ gives the $\lambda$-quantile, $Q$, and $T \in \mathbb{R}_+$ represents the temperature.

The optimal policy $\pi^{\ast}(U)$ is considered theoretically.
To achieve this, the expected optimality is given with $\pi^{\ast}(U)$ using the importance sampling technique~\cite{bishop2006pattern}.
\begin{align}
    &\mathbb{E}_{U_t^n \sim \pi(U_t; \theta)}[W_t^n]
    \nonumber \\
    =& \mathbb{E}_{U_t^n \sim \pi^{\ast}(U_t)}\left[ W_t^n \frac{\pi(U_t^n; \theta)}{\pi^{\ast}(U_t^n)} \right]
    =: W_t^e
    \label{eq:mpc_is}
\end{align}
The variance of $W_t^e$ should converge to the minimum with $\pi^{\ast}(U)$, hence, its likelihood can be derived~\cite{botev2013cross}.
\begin{align}
    \pi^{\ast}(U_t^n) = \frac{W_t^n}{W_t^e} \pi(U_t^n; \theta)
    \label{eq:mpc_opt_pi}
\end{align}

Although it is possible to theoretically derive $\pi^{\ast}(U_t)$ as mentioned above, it is difficult to use $\pi^{\ast}(U_t)$ as it is for MPC because $U_t^{\ast}$ cannot be obtained for practical use unless the mean, median or mode of $\pi^{\ast}(U_t)$ are known.
Hence, optimizing $\pi(U_t; \theta)$ (more specifically, $\theta$) in order to be closer to $\pi^{\ast}(U_t)$ by minimizing the following FKL divergence~\cite{botev2013cross}.
\begin{align}
    \theta^{\ast} &= \arg\min_{\theta^{\prime}} \mathrm{KL}(\pi^{\ast}(U_t) || \pi(U_t; \theta^{\prime}))
    \nonumber \\
    &= \arg\min_{\theta^{\prime}} \mathbb{E}_{U_t^n \sim \pi(U_t; \theta)}\left[ \frac{W_t^n}{W_t^e} \ln \frac{\pi^{\ast}(U_t^n)}{\pi(U_t^n; \theta^{\prime})} \right]
    \nonumber \\
    &= \arg\min_{\theta^{\prime}} \mathbb{E}_{U_t^n \sim \pi(U_t; \theta)}[ -W_t^n \ln \pi(U_t^n; \theta^{\prime}) ]
    \label{eq:mpc_fkl}
\end{align}
where
\begin{align*}
    \mathrm{KL}(p_1(x) || p_2(x)) = \mathrm{E}_{x \sim p_1(x)}\left[p_1(x)\ln \frac{p_1(x)}{p_2(x)} \right]
\end{align*}
with the two probabilities $p_1$ and $p_2$.
The optimization variable is represented by $\theta^{\prime}$, which is distinguished from $\theta$ for sampling the candidates.
The term not involved in $\theta^{\prime}$ is excluded, and $W_t^e$ is ignored by assuming that $w$ is appropriately designed so that $W_t^e = \mathrm{const.}$.
The Monte Carlo method approximates the remaining expectations with $N$ candidates. The sampling-based MPC optimizes $\theta$ using the maximum likelihood estimation, weighted by $W_t^n \in \mathbb{R}_{+}$ for every candidate $n$, $n=1,\ldots,N$.
Note that the larger value of $N$ is, the lower approximation error.

However, achieving this optimization in a single iteration may not be possible since there may be cases where none of the candidates cannot find sufficient optimality.
After optimizing $\theta$, the candidates are sampled and evaluated again, and $\theta$ is iteratively optimized.
To avoid overfitting during the iterations caused by the approximation error owing to insufficient candidates, many implementations use a smoothing method to update $\theta$, as recommended in~\cite{botev2013cross}.
Specifically, the update of $\theta$ for the $i$-th iteration ($i = 1,2,\ldots$) is given as follows:
\begin{align}
    \theta_{i+1} = (1 - \alpha) \theta_i + \alpha \theta^{\ast}
    \label{eq:mpc_smooth}
\end{align}
where $\alpha \in (0, 1]$ is the step size, and the smoothing can be interpreted as an update by the gradient method for eq.~\eqref{eq:mpc_fkl}.
Note that $\theta_{1}$ of the normal distribution is typically modeled as a standard normal distribution with $\mu=0$ and $\sigma=I$.

\subsection{Open issues}

\begin{figure}[tb]
    \centering
    \includegraphics[keepaspectratio=true,width=0.96\linewidth]{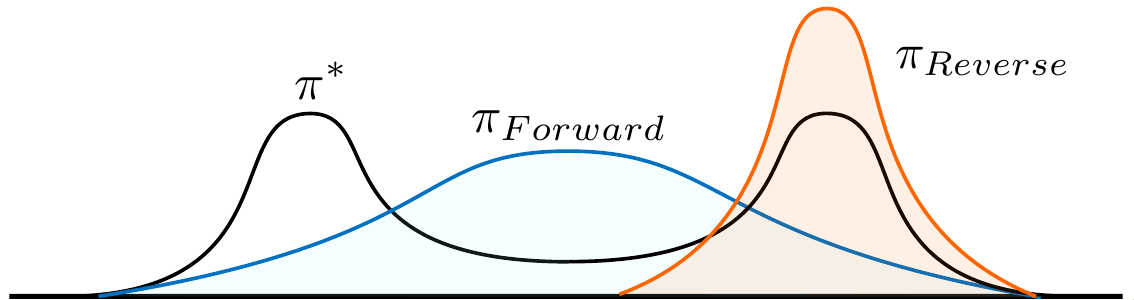}
    \caption{Convergence properties of FKL and RKL divergences.
    When minimizing FKL, $\pi$ obtains the mean of $\pi^\ast$, so-called mass-covering property.
    When minimizing RKL, $\pi$ obtains one of the modes of $\pi^\ast$, so-called mode-seeking property.
    }
    \label{fig:diff_fkl_rkl}
\end{figure}

Sampling-based MPC requires a sufficient number of candidates for stable updates, and obtains a truly optimal policy.
This is a major challenge for robot control, as it requires a high degree of real-time processing.
To optimize the conventional MPC, FKL divergence between the optimal policy and the policy to be optimized serves as the basis for optimization.
Depending on whether the target side of the two probabilities is given as the first or second argument, asymmetry can arise.
This asymmetry is widely known to alter the results of optimization obtained~\cite{minka2005divergence}, as shown in Fig.~\ref{fig:diff_fkl_rkl}.

Specifically, with $\pi^{\ast}$ the target and $\pi$ the optimization variable, the minimization of FKL divergence, $\mathrm{KL}(\pi^{\ast}(U) || \pi(U))$, includes $-\pi^{\ast}(U) \ln \pi(U)$ in the minimization objective.
If there exists $U$ that satisfies $\pi^{\ast}(U) \neq 0$ and $\pi(U) \simeq 0$, $-\pi^{\ast}(U) \ln \pi(U) \to \infty$.
For FKL divergence, $\pi^{\ast}(U) \neq 0$ and $\pi(U) \neq 0$ must both be satisfied for any $U$, resulting in $\pi$ hacing ``mass-covering,'' covering the entire target distribution.
In contrast, minimizing the RKL divergence, $\mathrm{KL}(\pi(U) || \pi^{\ast}(U))$ involves $-\pi(U) \ln \pi^{\ast}(U)$ in the objective function.
This term converges to zero when $\pi(U) \to 0$.
Therefore, RKL does not need to update $\pi$ to include the whole region with $\pi^{\ast}(U) \neq 0$, and can be minimized by fitting only one mode, as termed ``mode-seeking'' property.

Although mass-covering property with FKL divergence is effective for MPC to obtain a globally optimal solution, it can decrease computational efficiency since it requires more samples for evaluation.
When it comes to real-time control, if the time limit is exceeded, the iteration is stopped and the policy (and action) that is still being optimized is used.
This may result in a solution that is far from the globally optimal one in practice.
Hence, for practical purposes, it is preferable to achieve a degree of convergence in early iterations, even if it is a locally optimal solution (theoretically introduced in Section~\ref{sec:base}).

In addition, the introduction of smoothing, also known as gradient-based update, which stabilized the update of $\theta$, causes a trade-off between stability and efficiency with the step size.
Theoretically, a simple gradient method would result in only first-order convergence at the maximum.
While second-order convergence is ideally available by making full use of the momentum of the updates, as in NAG~\cite{nesterov1983method}, it has been reported in recent literature that acceleration is susceptible to gradient noise~\cite{cohen2018acceleration}.
The sensitivity to noise can significantly affect the (pseudo) gradient in MPC.
The objective function changes every iteration, and with fewer the number of candidates, approximation errors make it noisier.
Hence, it is desirable to introduce an acceleration method that takes this problem into account as heuristically proposed in Section~\ref{sec:impl}.

\section{Basic derivation}
\label{sec:base}

\subsection{Minimization of reverse Kullback-Leibler divergence}

To promote efficient convergence to a locally optimal solution, this study proposed a novel MPC by minimizing of RKL divergence as below.
\begin{align}
    \theta^{\ast} &= \arg\min_{\theta^{\prime}} \mathrm{KL}(\pi(U_t; \theta^{\prime}) || \pi^{\ast}(U_t))
    \nonumber \\
    &= \arg\min_{\theta^{\prime}} \mathbb{E}_{U_t^n \sim \pi(U_t; \theta^{\prime})}\left[ \ln \frac{W_t^e \pi(U_t^n; \theta^{\prime})}{W_t^n \pi(U_t; \theta)} \right]
    \nonumber \\
    &= \arg\min_{\theta^{\prime}} \mathbb{E}_{U_t^n \sim \pi(U_t; \theta^{\prime})}\left[ \ln \frac{\pi(U_t^n; \theta^{\prime})}{\pi(U_t; \theta)} \right]
    \nonumber \\
    &+ \mathbb{E}_{U_t^n \sim \pi(U_t; \theta)}\left[ \frac{\pi(U_t^n; \theta^{\prime})}{\pi(U_t^n; \theta)} (\ln W_t^e - \ln W_t^n) \right]
    \nonumber \\
    &= \arg\min_{\theta^{\prime}} \mathrm{KL}(\pi(U_t; \theta^{\prime}) || \pi(U_t; \theta))
    \nonumber \\
    &+ \mathbb{E}_{U_t^n \sim \pi(U_t; \theta)}\left[ \frac{\pi(U_t^n; \theta^{\prime})}{\pi(U_t^n; \theta)} (\ln W_t^e - \ln W_t^n) \right]
    \nonumber \\
    &= \arg\min_{\theta^{\prime}} \mathrm{KL}(\pi(U_t; \theta^{\prime}) || \pi(U_t; \theta))
    \nonumber \\
    &+ \mathbb{E}_{U_t^n \sim \pi(U_t; \theta)}\left[ - \frac{\pi(U_t^n; \theta^{\prime})}{\pi(U_t^n; \theta)} \ln W_t^n \right]
    \label{eq:mpc_rkl}
\end{align}
where the fourth line with $\mathbb{E}_{U_t^n \sim \pi(U_t; \theta)}[\cdot]$ is derived using the importance sampling, and in the derivation of the last line, $W_t^e$ is again assumed to be negligible as $W_t^e = \mathrm{const.}$ by appropriate $w$ design as in eq.~\eqref{eq:mpc_fkl}.
As a result, the first term regularizes to prevent $\theta^{\ast}$ from deviating from $\theta$, and the second term can be interpreted as the main objective.

The following can be derived by considering the gradient of the second term w.r.t. $\theta^{\prime}$.
\begin{align}
    &\nabla_{\theta^{\prime}} \mathbb{E}_{U_t^n \sim \pi(U_t; \theta)}\left[ - \frac{\pi(U_t^n; \theta^{\prime})}{\pi(U_t^n; \theta)} \ln W_t^n \right]
    \nonumber \\
    =& \mathbb{E}_{U_t^n \sim \pi(U_t; \theta)}\left[ - \frac{\pi(U_t^n; \theta^{\prime})}{\pi(U_t^n; \theta)} \ln W_t^n \nabla_{\theta^{\prime}} \ln \pi(U_t^n; \theta^{\prime}) \right]
    \nonumber \\
    \simeq& \mathbb{E}_{U_t^n \sim \pi(U_t; \theta)}\left[ - \ln W_t^n \nabla_{\theta^{\prime}} \ln \pi(U_t^n; \theta^{\prime}) \right]
    \label{eq:mpc_rkl_grad}
\end{align}
where $\nabla x = x \nabla \ln x$ is utilized in the second line.
In addition, assuming the last approximation is obtained through $\theta \simeq \theta^{\prime}$ is satisfied by the regularization in eq.~\eqref{eq:mpc_rkl}.
As a result, this is also the weighted maximum likelihood estimation with the weight $\ln W_t^n$ ($W_t^n \in \mathbb{R}_+$), which can be positive for $W_t^n > 1$ and negative for $W_t^n < 1$.

In summary, the Reverse MPC, starting from RKL divergence, differs from conventional MPC starting from FKL divergence at two points.
\begin{enumerate}
    \item Updating $\theta$ is restricted by RKL divergence between the policy before and after updating.
    \item The weight assigned to each candidate $\ln W_t^n$ takes not only a non-negative but also negative value.
\end{enumerate}
However, owing to these differences, the closed-form solution cannot be obtained even if $\pi$ is modeled as normal distribution.

\subsection{Solution with mirror descent algorithm}

\begin{figure}[tb]
    \centering
    \includegraphics[keepaspectratio=true,width=0.96\linewidth]{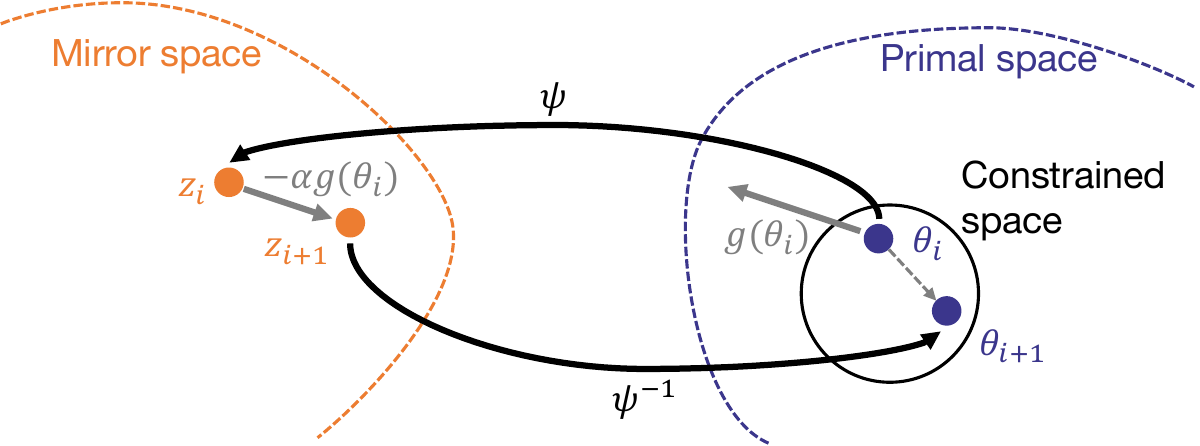}
    \caption{Overview of MD algorithm.
    The parameter in the primal space with constraints, $\theta$, is mapped into its mirror space without any constraints, $z$.
    The gradient w.r.t. $\theta$ is adopted to update $z$.
    The updated $z$ is again mapped into the primal space as the updated $\theta$.
    }
    \label{fig:framework_md}
\end{figure}

Together with eqs.~\eqref{eq:mpc_rkl} and~\eqref{eq:mpc_rkl_grad}, Reverse MPC can be formulated as a constrained minimization problem.
\begin{align}
    \theta_{i+1} &= \arg\min_{\theta^{\prime}} \mathbb{E}_{U_t^n \sim \pi(U_t; \theta_i)}\left[ - \ln W_t^n \ln \pi(U_t^n; \theta^{\prime}) \right]
    \nonumber \\
    \mathrm{s.t.} &\quad \ \mathrm{KL}(\pi(U; \theta^{\prime}) || \pi(U; \theta_i)) \leq \delta
    \label{eq:mpc_rkl_const}
\end{align}
where $\theta_{i, i+1}$ denotes the variable before and after optimization at $i$-th iteration, that is, $\theta$ in the above is replaced by $\theta_i$ for convenience.
The strictness of the constraint by RKL divergence is determined by $\delta$, although it is not specified explicitly in our method (see later).

Unlike eq.~\eqref{eq:mpc_rkl_const}, we need to numerically solve the constrained problem because there is no closed-form solution available eq.~\eqref{eq:mpc_fkl} (i.e. the weighted maximum likelihood estimation).
Furthermore, it is imperative that the numerical solution be sufficiently fast.
To discuss constrained optimization problems, one can use the MD algorithm~\cite{beck2003mirror}, which is a generalized gradient method.
This algorithm can handle RKL/FKL divergence as it is a type of Bregman divergence~\cite{bregman1967relaxation}.

Therefore, as can be seen in Fig.~\ref{fig:framework_md}, $\theta_i$ can be updated to $\theta_{i+1}$ using MD as follows:
\begin{align}
    \theta_{i+1} &= \psi^{-1}\left \{\psi(\theta_i) - \alpha g(\theta_i) \right \}
    \label{eq:mpc_rkl_md} \\
    g(\theta_i) &= \mathbb{E}_{U_t^n \sim \pi(U_t; \theta_i)}\left[ - \ln W_t^n \nabla_{\theta_i} \ln \pi(U_t^n; \theta_i) \right]
    \nonumber
\end{align}
where $\alpha$ denotes the step size as well as eq.~\eqref{eq:mpc_smooth} and is qualitatively proportional to $\delta$ since the updatable amount (amplified by $\alpha$) is implicitly defined by $\delta$.
The invertible function, $\psi$ (and $\psi^{-1}$), maps between the primal space to which $\theta$ belongs and the corresponding mirror space.
Note that $\psi$ is also the derivative of the given Bregman divergence (i.e. RKL divergence in this case).
If the update goal is to maximize the likelihood and RKL divergence is given, $\psi$ is the natural logarithm.
The above update formula is called exponential gradient descent.
It should be noted that the target of this update is a model parameter of the policy $\theta$.
In this study, we analytically derive $\psi_{\mu, \sigma}$ for location $\mu$ and scale $\sigma$ (i.e. $\theta = [\mu, \sigma]$) when modeling the policy with a diagonal normal distribution.
Note that with a closed-form KL divergence solution can also be used to model the policy.

Specifically, RKL divergence for the diagonal normal distribution has the closed-form solution as follows:
\begin{align}
    &\mathrm{KL}(\pi(U; \theta^{\prime}) || \pi(U; \theta_i))
    \nonumber \\
    =& \frac{1}{2} \sum_{k=1}^{|\mathcal{U}| \times H}
    \left \{ \ln \frac{\sigma_{i,k}^2}{\sigma_k^2} + \frac{\sigma_k^2}{\sigma_{i,k}^2} + \frac{(\mu_{i,k} - \mu_k)^2}{\sigma_{i,k}^2} - 1 \right \}
    \label{eq:kl_normal}
\end{align}
where the subscript $k$ denotes $k$-th dimension component in each parameter.
The derivative of each component is derived in a straightforward manner.
\begin{align}
    \begin{split}
        \psi_{\mu_k}(\theta^{\prime}) &= \nabla_{\mu_k}\mathrm{KL}(\pi(U; \theta^{\prime}) || \pi(U; \theta_i)) = \frac{\mu_k}{\sigma_{i,k}^2}
        \\
        \psi_{\sigma_k}(\theta^{\prime}) &= \nabla_{\sigma_k}\mathrm{KL}(\pi(U; \theta^{\prime}) || \pi(U; \theta_i)) = \frac{\sigma_k}{\sigma_{i,k}^2} - \frac{1}{\sigma_k}
    \end{split}
    \label{eq:psi_normal}
\end{align}
In addition, its inverse $\psi^{-1}$ can also be derived as follows:
\begin{align}
    \begin{split}
        \psi^{-1}_{\mu_k}(z) &= \sigma_{i,k}^2 z_{\mu_k}
        \\
        \psi^{-1}_{\sigma_k}(z) &= \frac{1}{2} \left ( \sigma_{i,k}^2 z_{\sigma_k} + \sigma_{i,k} \sqrt{\sigma_{i,k}^2 z_{\sigma_k}^2 + 4} \right)
    \end{split}
    \label{eq:psi_inv_normal}
\end{align}
where $z = \psi(\theta^{\prime}) \in \mathbb{R}^{2 \times |\mathcal{U}| \times H}$, with the subscript indicating the corresponding component.
Precisely, $\theta_i$ can be updated to $\theta_{i+1}$ using MD by substituting eqs.~\eqref{eq:psi_normal} and~\eqref{eq:psi_inv_normal} into eq.~\eqref{eq:mpc_rkl_md}.
Note that $\psi^{-1}$ is mapped to whole/non-negative real space with respect to $\mu$/$\sigma$, respectively.
This is particularly true because of $\sigma_{i,k}^2 z_{\sigma_k} \leq \sigma_{i,k} \sqrt{\sigma_{i,k}^2 z_{\sigma_k}^2 + 4}$.
In addition, the obtained update rule is similar to eq.~\eqref{eq:mpc_fkl} combined with eq.~\eqref{eq:mpc_smooth}.
Therefore, it would be a reasonable extension of the conventional method.

\subsection{Design of weight}

For the above Reverse MPC, a specific weight function $w$ needs to be designed as $\ln W = w(J)$ from the beginning.
Since $\ln W$ is allowed to be negative, unlike non-negative $w$ in the conventional MPC illustrated in eq.~\eqref{eq:mpc_weight_fkl}.
Then, $w$ is allowed to map $J$ to the whole real space.

In this study, the design example is presented as a simple extension of the conventional method.
First, a negative ratio $\beta \in [0, 1]$ is introduced, which indicates how much samples with poor results are revealed.
With $\beta$, the weight of $n$-th candidate ($n=1,\ldots,N$) is designed as follows:
\begin{align}
    \ln W_t^n = w_1(J(x_t, U_t^n)) - w_{\beta}(-J(x_t, U_t^n))
    \label{eq:mpc_weight_rkl}
\end{align}
where $w_{x=1,\beta}(\cdot)$ normalizes eq.~\eqref{eq:mpc_weight_fkl} so that its summation is $x$, for example,
\begin{align}
    &w_x(J(x_t, U_t^n)) =
    \\
    &\begin{cases}
        \cfrac{\mathbb{I}(J(x_t, U_t^n) \leq \mathrm{Q}(\{J(x_t, U_t^n)\}_{n=1}^N; x\lambda))}{\lambda N} & \mathrm{CEM}
        \\
        x \cfrac{\exp(-J(x_t, U_t^n) / T)}{\sum_{m=1}^N \exp(-J(x_t, U_t^m) / T)} & \mathrm{MPPI}
    \end{cases}
    \nonumber
\end{align}
The simplest normalization is to multiply $x=1,\beta$, after normalizing by division by the sum.
Meanwhile, in the case of CEM, it would be more efficiently cheap to implement a tighter threshold for outputting nonzero weights by $\beta$ (i.e. $\beta \lambda$-quantile).
Regardless, $w_{x=1,\beta}(\cdot)$, which reflects the fact that Reverse MPC is derived from RKL divergence, must have the ability to update the policy exclusively to prevent sampling the candidates that could yield worthless results in the future.
In addition, if $\beta = 0$, Reverse MPC is mostly reverted to the conventional method, although MD is remained.

\section{Practical implementation}
\label{sec:impl}

\subsection{Avoidance of interference with positive/negative updates}

\begin{figure}[tb]
    \centering
    \includegraphics[keepaspectratio=true,width=0.96\linewidth]{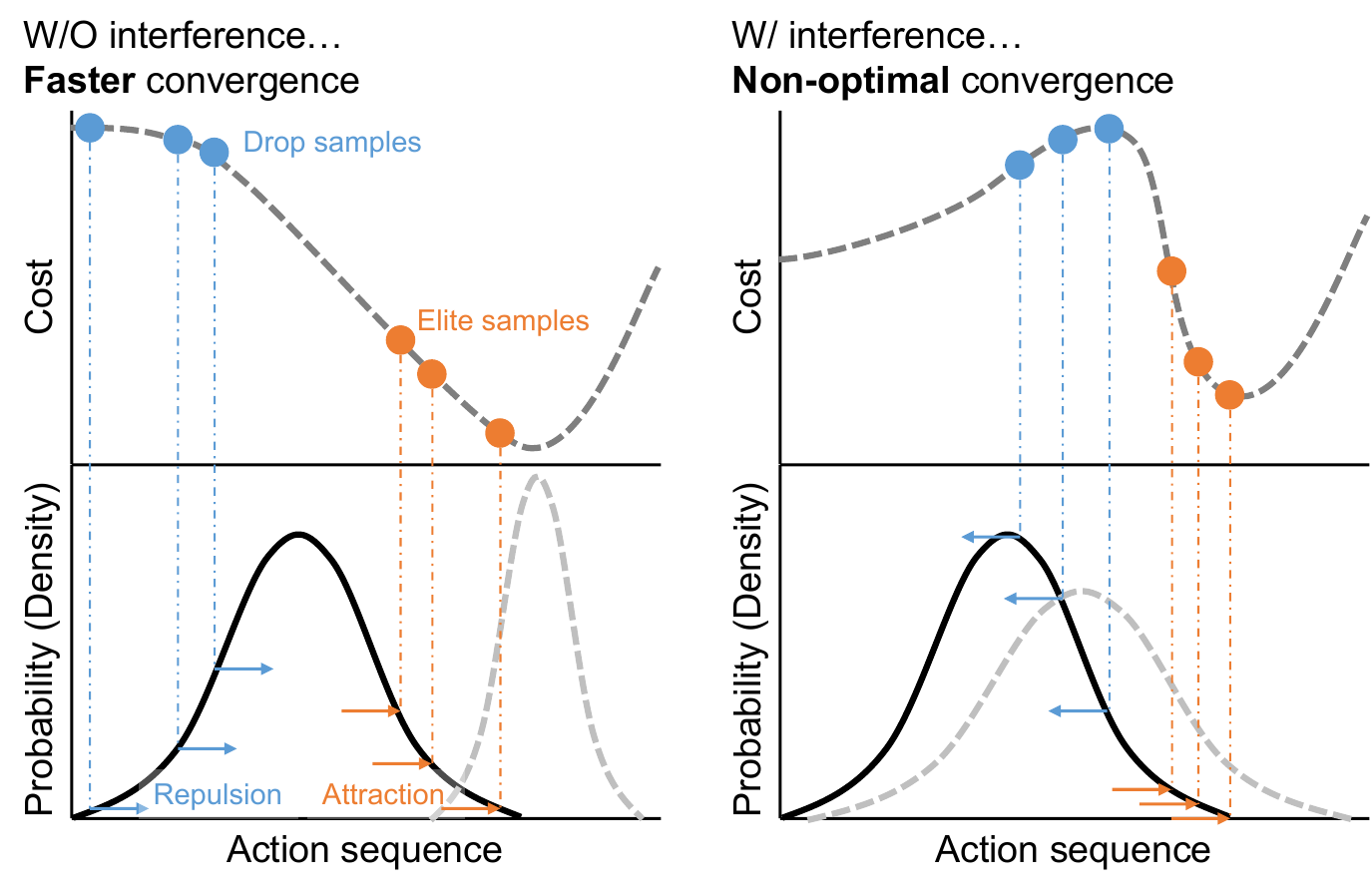}
    \caption{Interference with positive/negative updates.
    If the repulsion from the drop samples is in the same direction as the atraction to the elite samples, the policy can converge to the optimal one.
    If not, the repulsion conflicts with the atraction, preventing the desired updates of the policy.
    }
    \label{fig:problem_interference}
\end{figure}

\begin{figure}[tb]
    \centering
    \includegraphics[keepaspectratio=true,width=0.96\linewidth]{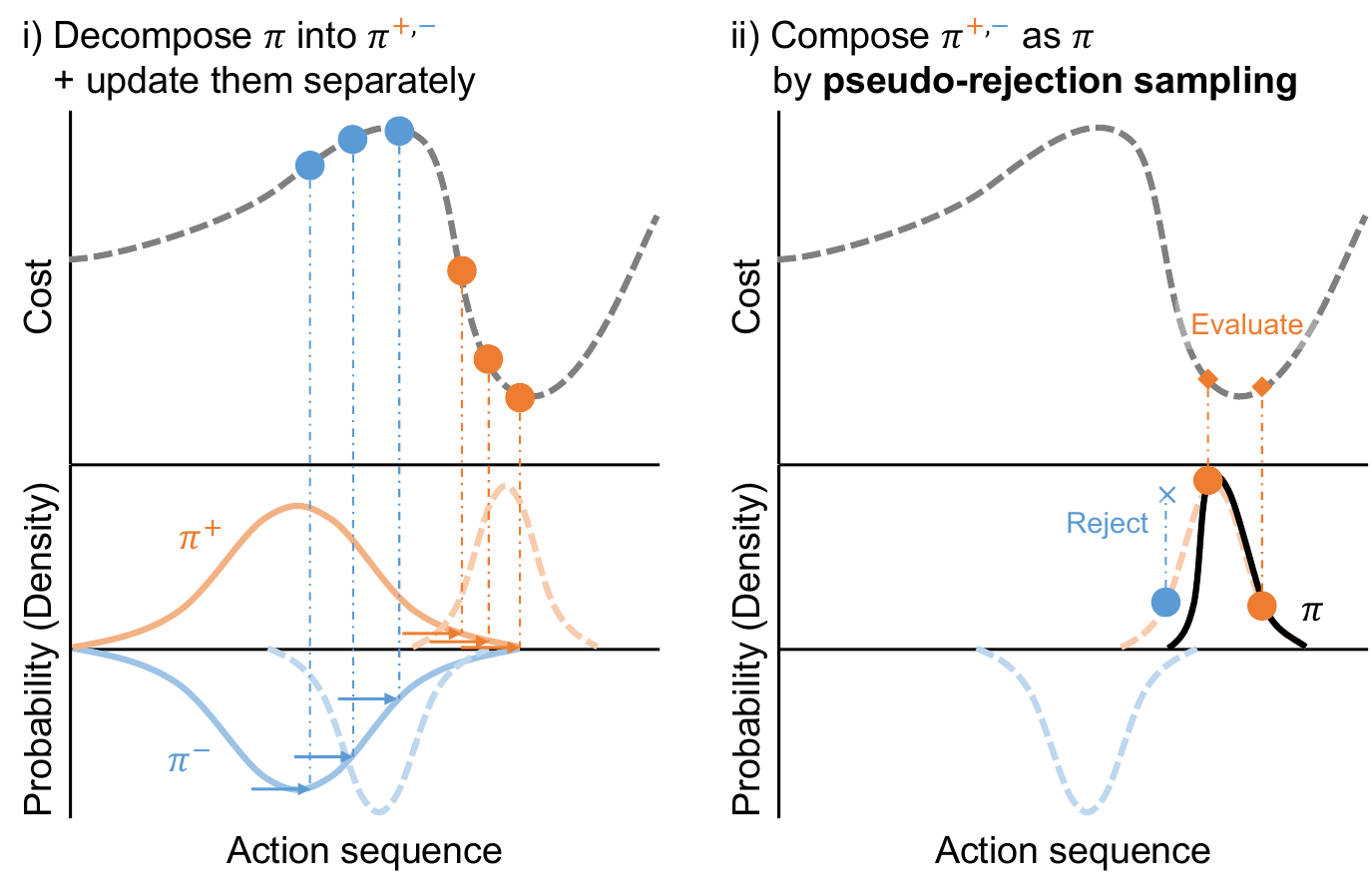}
    \caption{Proposal of Reject MPC.
    Two policies $\pi^{+,-}$ are updated to represent the elite and drop samples, respectively.
    They are composed as $\pi$ in a pseudo-rejection sampling manner.
    }
    \label{fig:framework_reject}
\end{figure}

Reverse MPC is a process that prioritizes good action sequences while avoiding those with poor results and is likely to improve convergence to a locally optimal solution. This process can generate non-fatal actions even if the iterations are terminated within a time limit.
However, this is only a theoretical expectation.
In numerical computation, a single scalar for the two types of weights (i.e. $w_1(J(x_t, U_t^n))$ and $w_{\beta}(-J(x_t, U_t^n))$ in eq.~\eqref{eq:mpc_weight_rkl}) can cause interference in the update.
As shown in Fig.~\ref{fig:problem_interference}, it is possible to accelerate convergence if the direction of negative updates away from the drop samples is consistent with positive updates approaching elite samples.
However, depending on the shape of the cost function and the candidates sampled, these update directions may counteract each other, resulting in a slower convergence rate.
Interference within positive updates can occur, especially when the task is multiobjective, but in Reverse MPC, the addition of negative updates makes this more pronounced.

By simply making the policy model more expressive, such as by using mixture distribution~\cite{okada2020variational} or normalizing flow~\cite{sacks2023learning}, it is likely that the updates will no longer be affected by interference.
However, this approach not only significantly increases the computational cost but may also make it impossible to solve MD analytically.
To address this issue, we suggest a practical implementation of i) policy decomposition for updating and ii) composition of the decomposed policies for sampling, as discussed in Fig.~\ref{fig:framework_reject}.
This approach, known as Reject MPC, aims to solve the capability of solving various tasks by avoiding interference while minimizing the increase in computational cost.

\subsubsection{Update of decomposed policies}

As the first step of this approach, the policy is divided into $\pi^+(U_t; \theta^+)$ and $\pi^-(U_t; \theta^-)$.
Each model parameters set $\theta^{+,-}$ (i.e. $\theta = [\theta^+, \theta^-]$) is prepared.
Both of $\theta^{+,-}$ are updated for the clusters $C^{+,-}$ with positive/negative $\ln W_t^n$ from $N$ candidates, respectively.
The following update formulas are given for the respective $\theta^{+,-}$.
\begin{align}
    \theta_{i+1}^{+,-} &= \psi_{+,-}^{-1} \left \{ \psi_{+,-}(\theta_i^{+,-}) - \alpha g(\theta^{+,-}) \right \}
    \label{eq:mpc_rkl_pn} \\
    g(\theta^{+,-}) &= - \frac{1}{|C^{+,-}|} \sum_{n \in C^{+,-}} \ln W_t^n
    \nonumber \\
    &\quad\quad\quad\quad\quad\,\times \nabla_{\theta_i^{+,-}} \ln \pi^{+,-}(U_t^n; \theta_i^{+,-})
    \nonumber
\end{align}
Note that $\psi_{+,-}$ (and $\psi_{+,-}^{-1}$) are based on the respective $\theta_i^{+,-}$, as shown in eqs.~\eqref{eq:psi_normal} and~\eqref{eq:psi_inv_normal}, if their corresponding policies are modeled as the diagonal normal distribution.
With this update, $\pi^+(U_t; \theta^+)$ generally reinforces the good action sequences as like the conventional method.
On the other hand, $\pi^-(U_t; \theta^-)$ reveals the distribution that generates the bad action sequences.

\subsubsection{Pseudo-rejection sampling from composed policy}

The action sequence $U_t$ generated from the policy $\pi(U_t; \theta)$ that should theoretically be obtained by Reverse MPC would be $U_t \sim \pi^+(U_t; \theta^+)$ and $U_t \not\sim \pi^-(U_t; \theta^-)$.
That is, the policy corresponding to $\pi(U_t; \theta)$ is obtained by composing $\pi^+(U_t; \theta^+)$ and $\pi^-(U_t; \theta^-)$ as follows:
\begin{align}
    \pi(U_t; \theta) &\propto \pi^+(U_t; \theta^+) \bar{\pi}^-(U_t; \theta)
    \label{eq:policy_compose} \\
    \bar{\pi}^-(U_t; \theta) &\propto \frac{1}{\pi^-(U_t; \theta^-) + \pi^+(\mu^-; \theta^+)}
    \label{eq:policy_not_bad}
\end{align}
where $\bar{\pi}^-(U_t; \theta)$ represents the complementary distribution of $\pi^-(U_t; \theta^-)$ and is designed in an ad-hoc manner.
The reason for adding $\pi^+(\mu^-; \theta^+)$ to the denominator is to ensure correct sampling even when $\pi^+(U_t; \theta^+)$ and $\pi^-(U_t; \theta^-)$ overlap (see Appendix~\ref{app:ablation}).

Although, $\pi(U_t; \theta)$ can be inferred as eq.~\eqref{eq:policy_compose}, it is not directly possible to sample the candidates of the action sequence from it.
Therefore, with reference to rejection sampling~\cite{bishop2006pattern}, the action sequence $U_t^{\tilde{n}}$ sampled from $\pi^+(U_t; \theta^+)$ is evaluated using the following rejection criterion to determine whether to reject or accept it.
\begin{align}
    \frac{\pi(U_t^{\tilde{n}}; \theta)}{Z \pi^+(U_t^{\tilde{n}}; \theta^+)} \propto \bar{\pi}^-(U_t^{\tilde{n}}; \theta)
    \label{eq:reject_weight}
\end{align}
where $Z$ denotes the normalization constant of $\pi(U_t^{\tilde{n}}; \theta)$.
The smaller $\bar{\pi}^-(U_t^{\tilde{n}}; \theta)$, the more preferentially it is rejected.
In standard rejection sampling, each sample is independently evaluated with $Z$ (found a priori or adaptively) for rejection or acceptance.
This process is repeated until the required number of samples ($N$ in this case) is accepted.
Performing repeated computations can negatively impact real-time performance, because the amount of time required to complete these computations can vary.
Therefore, in Reject MPC, after sampling $\{U_t^{\tilde{n}}\}_{\tilde{n}=1}^{\tilde{N}}$ with $\tilde{N} \gg N$ first, the relatively unsuitable ones for $\pi(U_t; \theta)$ are rejected, selecting $\{U_t^n\}_{n=1}^N \subset \{U_t^{\tilde{n}}\}_{\tilde{n}=1}^{\tilde{N}}$ in one batch process with the constant computational time depending on $\tilde{N}$.
The selection probability of each candidate can be defined without finding $Z$ as follows:
\begin{align}
    p^{\tilde{n}} = \frac{\bar{\pi}^-(U_t^{\tilde{n}}; \theta)}{\sum_{k=1}^{\tilde{N}} \bar{\pi}^-(U_t^k; \theta)}
    \label{eq:reject_prob}
\end{align}

\subsection{Acceleration of convergence}

MD for its optimization in reverse and Reject MPC, but because it is classified as a first-order gradient method.
Its convergence is still limited to first-order and is not sufficiently fast.
Therefore, latest NAG in MD, AGD+~\cite{cohen2018acceleration}, is utilized to improve the convergence speed while modifying it for the dynamic mirror space.
However, AGD+ has pointed out the sensitivity to noise in the gradient wherein MPC easily causes such noise.
This is because of the approximation error of Monte Carlo method in MPC, which would bias the candidates sampled at each iteration differently.
To compensate for the varying level of bias, a heuristic is designed that estimated how noisy the gradient is.
This heuristic is then used to adjust the step size accordingly.

\subsubsection{Modified AGD+ for dynamic mirror space}

AGD+, as mentioned above, is the latest algorithm of NAG for MD~\cite{cohen2018acceleration}.
Specifically, AGD+ provides the following update rule.
\begin{align}
    \begin{split}
        z_i &= z_{i-1} - a_i g(\theta_i)
        \\
        y_i &= \frac{A_{i-1}}{A_i} y_{i-1} + \frac{a_i}{A_i} \psi^{-1}(z_i)
        \\
        \theta_{i+1} &= \frac{A_i}{A_{i+1}} y_i + \frac{a_{i+1}}{A_{i+1}} \psi^{-1}(z_i)
    \end{split}
    \label{eq:agd_origin}
\end{align}
where $g(\theta_i)$ denotes the gradient w.r.t. $\theta_i$.
In addition, $a_i = \alpha i$ and $A_i = \sum_{k=1}^i a_k$, respectively.

Since many variables (i.e. $z$, $y$, and $\theta$) need to be stored.
It is difficult to understand how the acceleration occurs, AGD+ can be converted to the rule with the momentum that is typical of NAG.
\begin{align}
    \begin{split}
        z_i &= z_{i-1} - a_i g(\theta_i)
        \\
        \theta_{i+1} &= \frac{A_i}{A_{i+1}} \theta_i + \frac{a_{i+1}}{A_{i+1}} \psi^{-1}(z_i)
        \\
        &+ \frac{a_i}{A_{i+1}} \left \{ \psi^{-1}(z_i) - \psi^{-1}(z_{i-1}) \right \}
    \end{split}
    \label{eq:agd_momentum}
\end{align}
where the initial $\theta_1$ is given and $z_1 = \psi(\theta_1)$.
The updated parameters are obtained by interpolating between the new values and previous parameters while incorporating a momentum term.
In this form, $y$ vanishes, thereby reducing the computational cost.
It can be easily confirmed in eq.~\eqref{eq:agd_origin}, which is equivalent to eq.~\eqref{eq:agd_momentum}, that $\theta$ is always placed in its constrained primal space.

It is important to note that the mirror space in the proposed Reverse (and Reject) MPC is dynamic for each iteration according to $\theta_i$.
Namely, $\psi^{-1}(z_i)$ is $\theta_i$-dependent, whereas $\psi^{-1}(z_{i-1})$ is $\theta_{i-1}$-dependent.
In addition, $z_{i-1}$ used to update $z_i$ is actually in mirror space, depending on $\theta_{i-1}$.
Hence, it should be remapped to the $\theta_i$-dependent mirror space.
Considerably, the proposed acceleration for Reverse (and Reject) MPC, so-called Accel MPC, modifies the update rule as follows:
\begin{align}
    \begin{split}
        \tilde{\theta}_i &= \psi^{-1} \left \{ \psi(\tilde{\theta}_{i-1}) - a_i g(\theta_i) \right \}
        \\
        \theta_{i+1} &= \frac{A_i}{A_{i+1}} \theta_i + \frac{a_{i+1}}{A_{i+1}} \tilde{\theta}_i
        + \frac{a_i}{A_{i+1}} \left \{ \tilde{\theta}_i - \tilde{\theta}_{i-1} \right \}
    \end{split}
    \label{eq:agd_mpc}
\end{align}
where $\psi$ (and $\psi^{-1}$) is dependent on $\theta_i$ only.
We introduce $\tilde{\theta}_i$ in place of $z_i$, whose initial value is $\tilde{\theta}_1 = \theta_1$.
When considering the mirror space, we only need to store $\theta_i$ and $\tilde{\theta}_{i-1}$ with this updated rule.

\subsubsection{Adaptive step size for gradient noise}

To adjust the step size for making the acceleration robust to gradient noise~\cite{cohen2018acceleration}, the strength of gradient noise is first estimated from the sample statistics.
The scale of samples should be proportional to noise.
If the distribution of samples is regarded to be non-Gaussian (or non-uniform), the effect of noise would be strongly evident.

\begin{figure}[tb]
    \centering
    \includegraphics[keepaspectratio=true,width=0.96\linewidth]{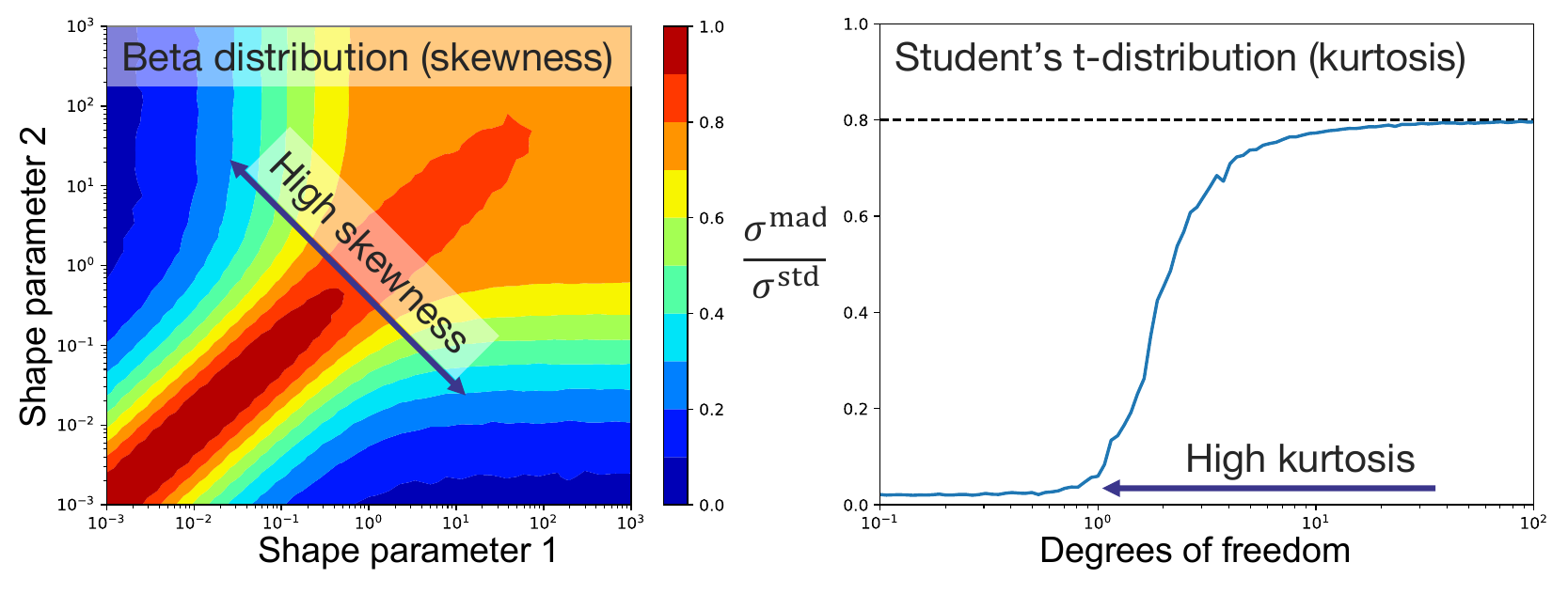}
    \caption{Evaluation of skewness and kurtosis by $\sigma^{\mathrm{mad}} / \sigma^{\mathrm{std}}$.
    On the left, the ratio decreases if the skewness increases.
    The kurtosis has the same tendency, inverse proportion to the ratio, as shown on the right.
    }
    \label{fig:diff_std_mad}
\end{figure}

Mathematically, the ratio of the sample standard deviation in each iteration, $\sigma_i^{\mathrm{std}}$, and the approximate maximum of the scale, $\sigma^{\mathrm{max}}$, can simply evaluate the former in a generalized manner.
For the latter, however, it is difficult to normalize the sample skewness and kurtosis based on a normal distribution.
Instead, the mean absolute deviation for each iteration is introduced, $\sigma_i^{\mathrm{mad}}$, to represent such higher sample statistics according to the inequality $\sigma_i^{\mathrm{std}} \geq \sigma_i^{\mathrm{mad}}$.
The equality is valid when the distribution of samples is uniform, and if the distribution is normal, $\sigma_i^{\mathrm{mad}} / \sigma_i^{\mathrm{std}} = 0.8$ can be obtained analytically.
The larger the skewness (more precisely, its absolute value) and/or the larger the kurtosis (i.e.
the heavier the tail), the larger the difference between $\sigma_i^{\mathrm{std}}$ and $\sigma_i^{\mathrm{mad}}$ will be (see Fig.~\ref{fig:diff_std_mad}).

In summary, the strength of noise is represented as follows:
\begin{align}
    s_i = \left (1 - \frac{\sigma_i^{\mathrm{mad}}}{\sigma_i^{\mathrm{std}}} \right)
    \times \frac{\sigma_i^{\mathrm{std}}}{\sigma^{\mathrm{max}}}
    \label{eq:noise_strength}
\end{align}
If the noise is ignorable, $s_i \to 0$; otherwise, $s_i \to 1$ (if $\sigma^{\mathrm{max}}$ is appropriately given).

Finally, the adaptive step size is designed with $s_i$.
\begin{align}
    a_{i+1} = a_i + \alpha \frac{1}{1 + 5\gamma s_i}
    \label{eq:agd_slowdown}
\end{align}
where $\gamma \geq 0$ determines the ease of slowdown according to the noise, and $a_1$ is given to be $\alpha$.
The reason for setting $5\gamma$ is that the first term of $s_i$ is $0.2$ within the normal distribution, wherein it can be canceled by multiplying $5$.
It is unclear whether the calculated sample scales are actually based on the values (and the optimal $\gamma$).
Hence, in Appendix~\ref{app:ablation}, this question is verified through testing the values related to the gradient computation.

\section{Experiments}

\subsection{Conditions}

\begin{figure}[tb]
    \centering
    \includegraphics[keepaspectratio=true,width=0.96\linewidth]{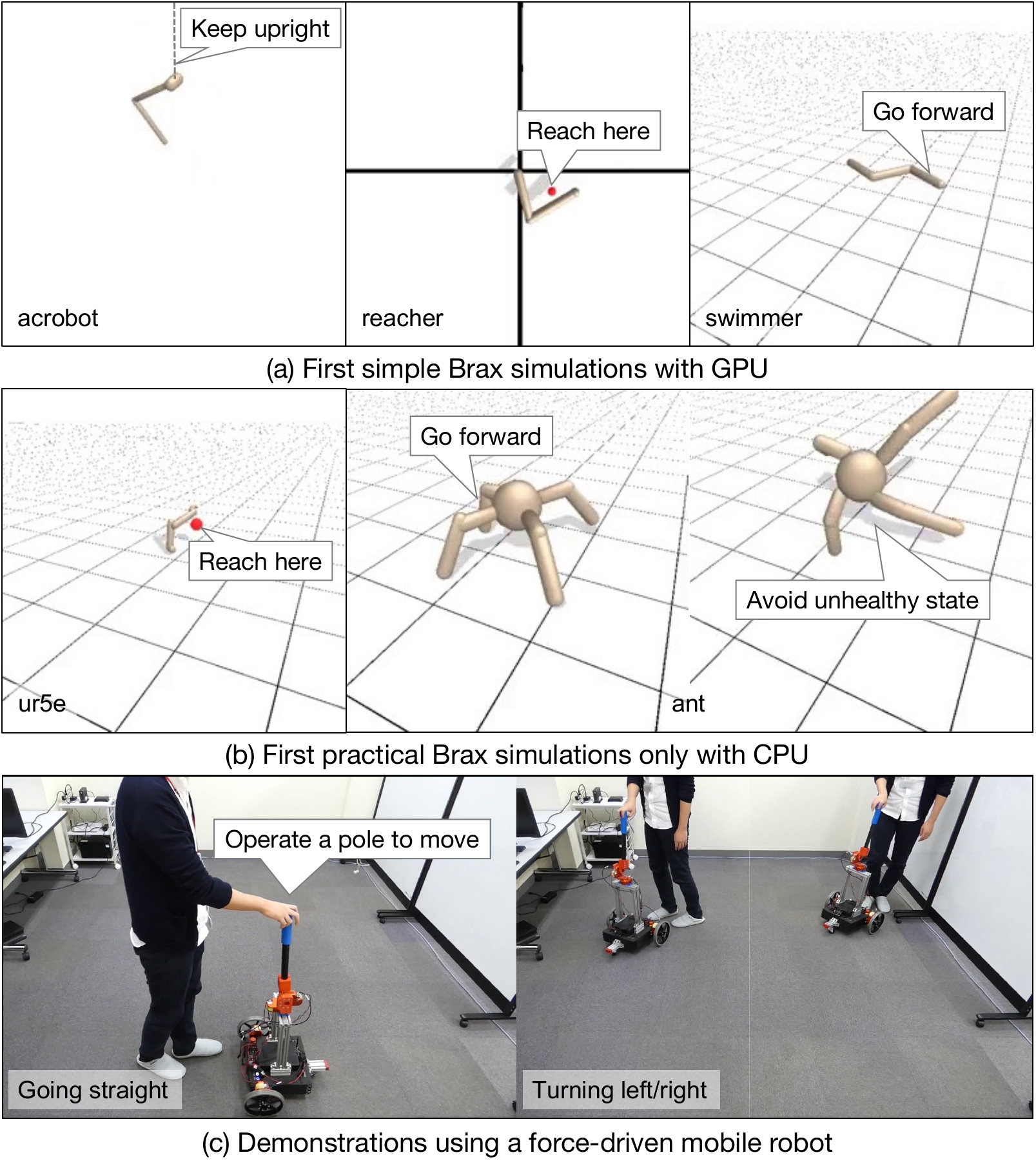}
    \caption{Tasks conducted.
    On the top, three toy problems are chosen to verify the proposed method without the acceleration.
    On the middle, two practical tasks with higher degrees of freedom are chosen to verify the acceleration.
    On the bottom, the real-world performance is demonstrated through the improvement of the operability.
    }
    \label{fig:task}
\end{figure}

\begin{table*}[tb]
    \caption{Parameter configuration:
    the tuples with three values are corresponding to the three experiments.
    }
    \label{tab:param}
    \centering
    \begin{tabular}{ccc}
        \hline\hline
        Symbol & Meaning & Value
        \\
        \hline
        $H$ & Horizon & $12$
        \\
        $N$ & \#Candidate & $\{4096, 32, 32\}$
        \\
        $(\mu_1, \sigma_1)$ & Prior policy & $(0, I)$
        \\
        $\alpha$ & Step size & $\{0.5, 0.05, 0.05\}$
        \\
        $\beta$ & Negative ratio & $1$
        \\
        $\tilde{N}$ & \#Candidate with rejection & $\{16N, 4N, 4N\}$
        \\
        $\gamma$ & Slowdown gain & $0.5$
        \\
        $\eta$ & Warm ratio (see Appendix~\ref{app:warm}) & $\{0, 0.25, 0.5\}$
        \\
        -- & Solver & \{CEM, MPPI, MPPI\}
        \\
        $\lambda$ & Quantile position (CEM~\cite{botev2013cross}) & $1$\%
        \\
        $T$ & Temperature (MPPI~\cite{williams2018information}) & $1$
        \\
        \hline\hline
    \end{tabular}
\end{table*}

The real-time control performance of the proposed method is verified step by step.
As a remark, the difference in convergence to the multimodal optimal policy, as shown in Fig.~\ref{fig:diff_fkl_rkl}, is presented in Appendix~\ref{app:toy}.
First, each method is statistically evaluated through benchmarks in Brax simulator~\cite{freeman2021brax} with or without a GPU.
For statistical evaluation, 100 trials with different initial states are tested for each solver and task.
After collecting the sum of the rewards (i.e. the negative cost) defined by each task as scores, the maximum and minimum scores for each task are found to normalize the scores in the range $[0, 1]$.

Then, the final method, Accel MPC, is applied to variable impedance control~\cite{buchli2011learning,ficuciello2015variable} of a force-driven mobile robot~\cite{kobayashi2022light} to demonstrate its operability.
Clearly, this demonstration exhibits stochastic behaviors because a user is integrated into the system.
Snapshots of the tasks conducted are listed in Fig.~\ref{fig:task}.
The details are summarized in Appendix~\ref{app:task}.

Each method is implemented using JAX~\cite{schoenholz2020jax}, which is a JIT-compilable library written in Python.
The given target systems with $f(x, u)$ and $C(x, u)$ are given.
That is, the simulation tasks utilize Brax simulator as well; and for the real robot, it is assumed that the robot moves perfectly, according to the commands, and the externally-applied force is constant as the recent average.
Note that in the simulation tasks, the number of internal steps of Brax simulator for MPC iterations is coarsened by $1/4$ to reduce the computational cost.
MPC iterations are maximized not to exceed the task control period (10~Hz/0.1~sec for simulation and 50~Hz/0.02~sec for real) for any method.
The hyperparameters of the methods are roughly tuned empirically and are listed in Table~\ref{tab:param}.
A set of three values in the table indicated that different values were used for the following three experiments, respectively (which were mainly due to the presence/absence of a GPU).

\subsection{Simulation results for Reverse and Reject MPCs}
\label{subsec:sim1}

\begin{figure}[tb]
    \centering
    \includegraphics[keepaspectratio=true,width=0.96\linewidth]{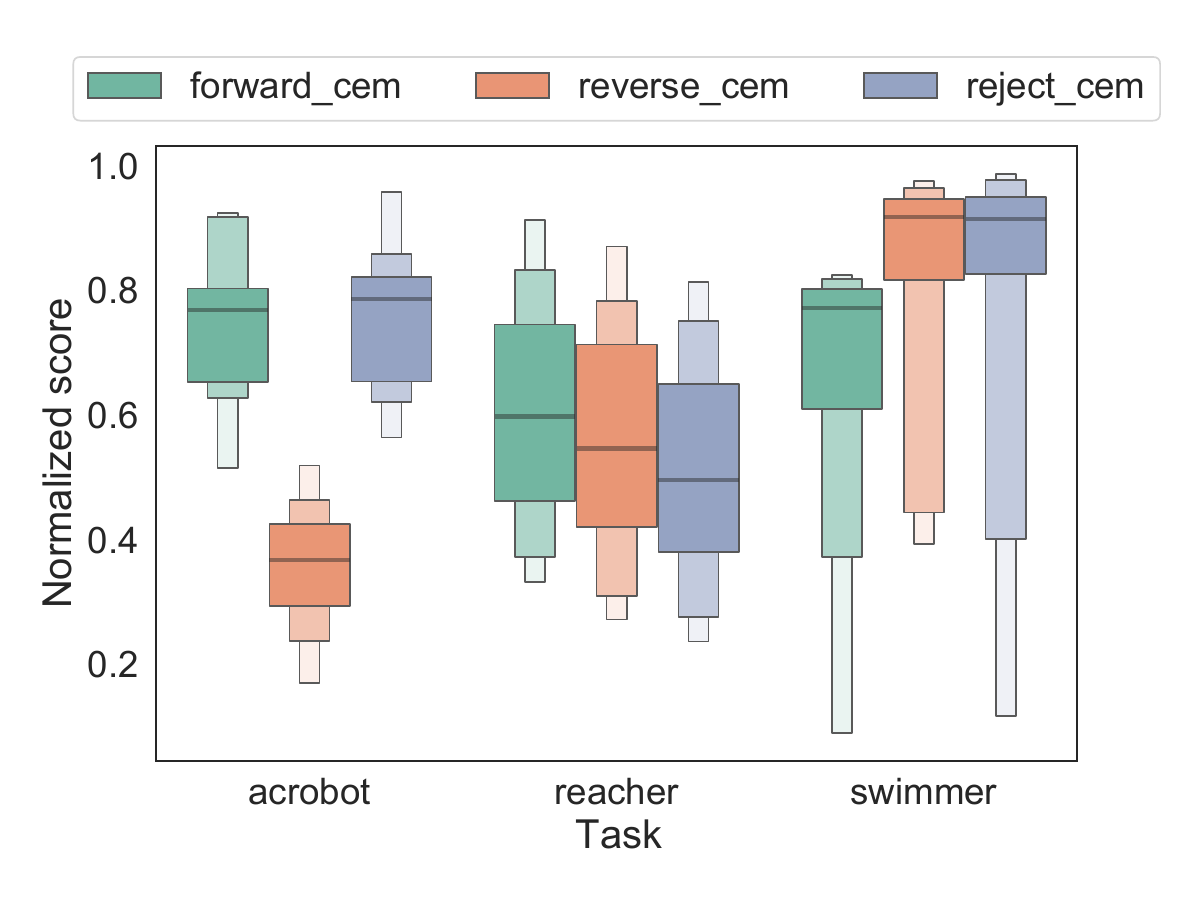}
    \caption{Scores for the simple tasks with GPU.
    Two baselines, \textit{forward\_cem} and \textit{reverse\_cem}, failed to solve one of the three tasks.
    The proposal, \textit{reject\_cem}, achieved the better generalizability than others.
    }
    \label{fig:sim_result_reject}
\end{figure}

\begin{figure}[tb]
    \centering
    \includegraphics[keepaspectratio=true,width=0.96\linewidth]{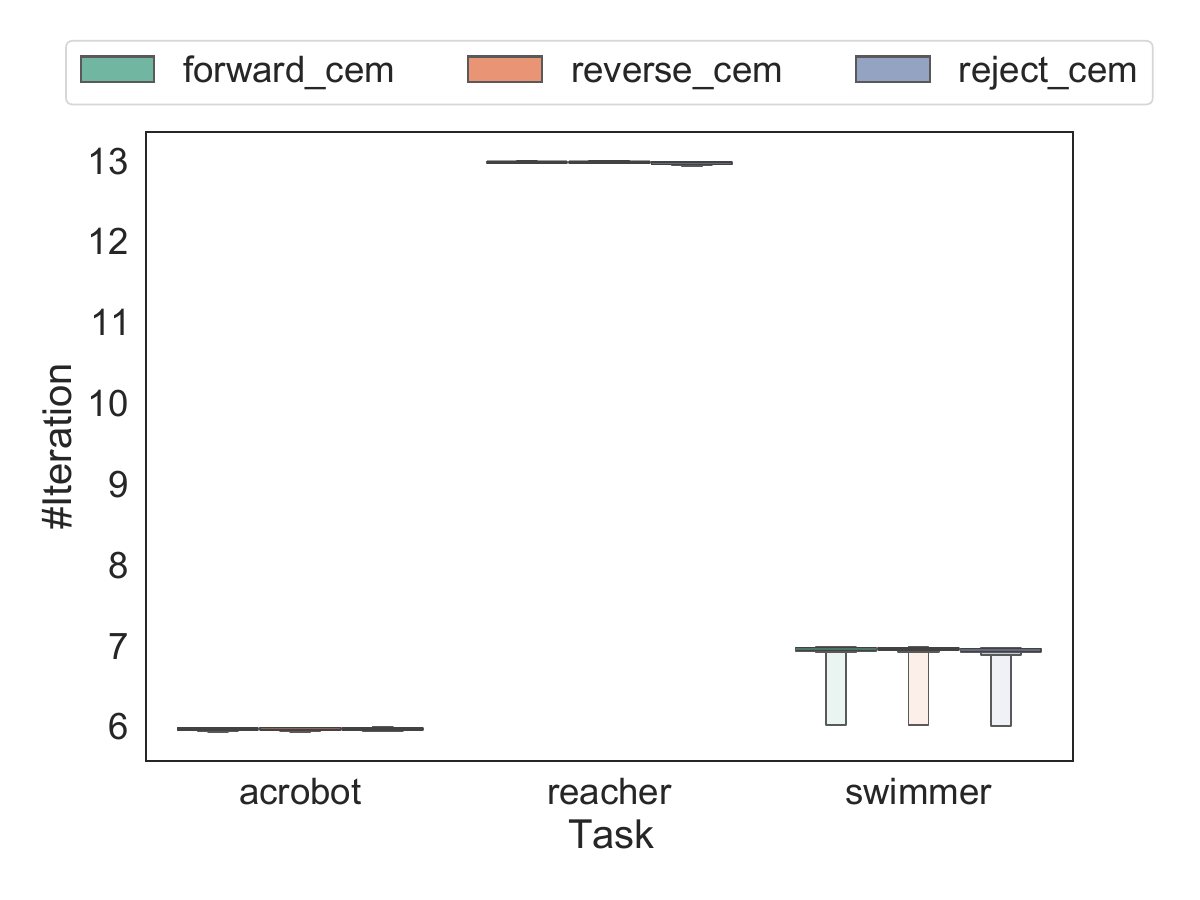}
    \caption{\#Iterations for the simple tasks with GPU.
    The proposed method achieved approximately the same number of iterations as the baselines with little increase in computational cost.
    }
    \label{fig:sim_iter_reject}
\end{figure}

As a first-stage simulation, the control performance on simple tasks is verified with a GPU.
Here, since sufficient candidates (i.e. $N=4096$) can be evaluated through the GPU, CEM~\cite{botev2013cross} is employed as the solver.
With it, \textit{forward\_cem} (the conventional CEM~\cite{botev2013cross}), \textit{reverse\_cem} (the theoretically proposed method), and \textit{reject\_cem} (the practically proposed method) are compared.
The results of testing \textit{acrobot}, \textit{reacher}, and \textit{swimmer} using these methods are shown in Figs.~\ref{fig:sim_result_reject} and~\ref{fig:sim_iter_reject}.

First, it was observed that the number of iterations remained constant for each task, indicating an insiginificant increase in the computational cost with the proposal when using the GPU.
As shown in Fig.~\ref{fig:sim_result_reject}, \textit{forward\_cem} and \textit{reverse\_cem} clearly deteriorated their control performance in specific tasks.
Apparently, \textit{forward\_cem} could not sufficiently optimize \textit{swimmer}, probably because of its relatively large action space, and; \textit{reverse\_cem} failed to acquire the (locally) optimal solution in \textit{acrobot} due to the interference with the positive/negative updates.
This is implied by the fact that \textit{reverse\_cem} achieved only the rotation motion, whereas the others succeeded in balancing on the top.
In contrast, \textit{reject\_cem} succeeded in all tasks, although the result of \textit{reacher} with \textit{reject\_cem} was slightly inferior to that of the others (with no significant difference).
This results suggested that the ad-hoc implementation of $\bar{\pi}^-(U_t; \theta)$ in eq.~\eqref{eq:policy_not_bad} should be improved in the near future.


\subsection{Simulation results for Accel MPC}
\label{subsec:sim2}

In the upcoming simulation, more complex tasks are tested using only a CPU.
The proposed acceleration method aims to improve convergence in cases where the conventional method fails to converge over time.
However, the benefits of GPU parallelization are not always available, as shown in the following real-robot demonstration.
In addition, they deal with discrete systems and/or costs, which gradient-based MPC cannot handle, and the impact of such discrete events on convergence is significant.

\begin{figure}[tb]
    \centering
    \includegraphics[keepaspectratio=true,width=0.96\linewidth]{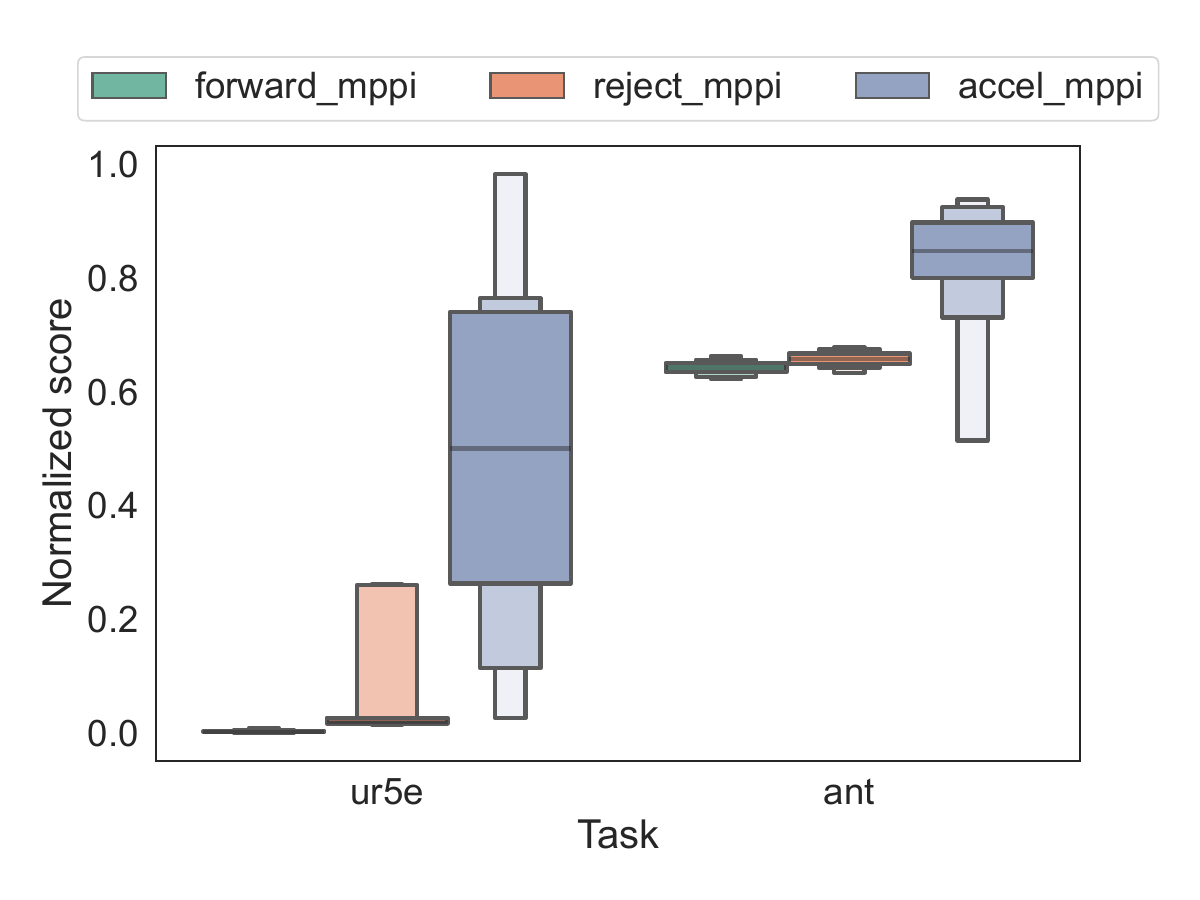}
    \caption{Scores for the complex tasks only with CPU.
    The baselines without the acceleration, \textit{forward\_mppi} and \textit{reject\_mppi}, failed to solve the tasks due to slow convergence.
    Only the proposed method with the acceleration, \textit{accel\_mppi}, could solve the tasks.
    }
    \label{fig:sim_result_accel}
\end{figure}

\begin{figure}[tb]
    \centering
    \includegraphics[keepaspectratio=true,width=0.96\linewidth]{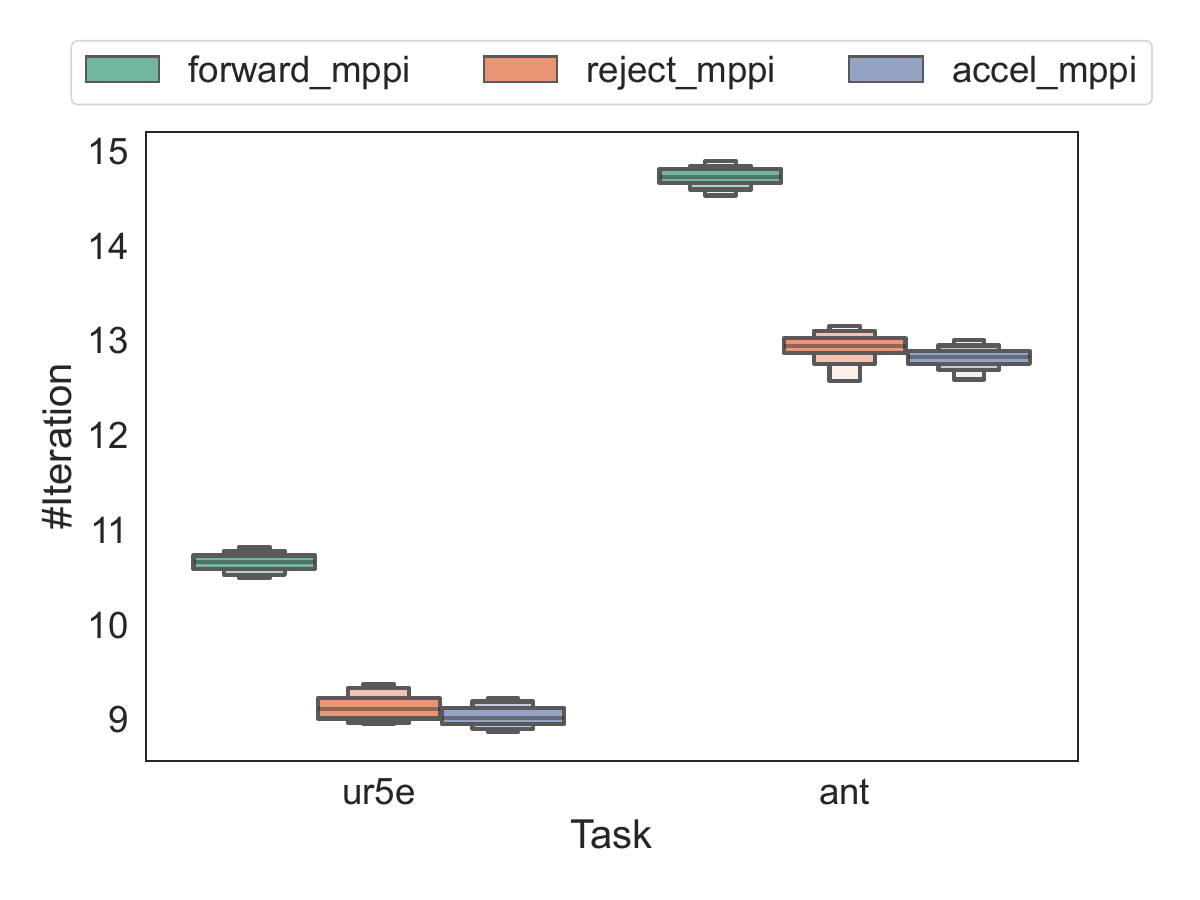}
    \caption{\#Iterations for the complex tasks only with CPU.
    Unlike the case with GPU, the case with CPU alone cannot suppress the increase in the computational cost by the pseudo-rejection sampling through parallelization, decreasing the number of iterations.
    }
    \label{fig:sim_iter_accel}
\end{figure}

According to the reduced number of candidates $N=32$ due to the poor parallel computation, the step size is set to $\alpha=0.05$ in order to stabilize the update.
Instead of reducing $\alpha$, the warm start (see in Appendix~\ref{app:warm}) is introduced to improve efficiency.
Furthermore, to make the best use of a small number of candidates, MPPI~\cite{williams2018information} is employed as the solver.
That is, \textit{forward\_mppi} (the conventional MPPI~\cite{williams2018information}), \textit{reject\_mppi} (the best method in the first simulation), and \textit{accel\_mppi} (the method accelerates \textit{reject\_mppi}) are compared.
The results of testing \textit{ur5e} and \textit{ant} with these methods are shown in Figs.~\ref{fig:sim_result_accel} and~\ref{fig:sim_iter_accel}.

The number of iterations was maintained at the same level (or higher) as in the first simulation, despite the increased complexity of the task.
This is because the CPU showed high single computational performance.
However, unlike previous simulations using GPU (see Fig.~\ref{fig:sim_iter_accel}), the number of iterations of the proposed method decreased when only CPU was used.
In fact, because of the proposed method, pseudo-rejection sampling requires parallelization, which is insufficient for the CPU.

It is found that \textit{forward\_mppi} perfectly failed \textit{ur5e} while \textit{reject\_mppi} was better than that but still statistically bad result.
In comparison, \textit{accel\_mppi} certainly accelerated the convergence to acquire reaching behavior.
In \textit{ant}, \textit{forward\_mppi} and \textit{reject\_mppi} obtained stable scores, but the motion with that scores was to stay in place, not walking at all.
In contrast, \textit{accel\_mppi} obtained a walking motion and a statistically high score, although it sometimes caused an unhealthy state and was terminated with a low score.
It should be noted that, with $\alpha=0.5$, the same value as in the first simulation, \textit{forward\_mppi} and \textit{reject\_mppi} sometimes succeeded in both tasks, but their scores were unstable.


\subsection{Demonstrations with force-driven mobile robot}
\label{subsec:exp}

\begin{figure}[tb]
    \centering
    \includegraphics[keepaspectratio=true,width=0.9\linewidth]{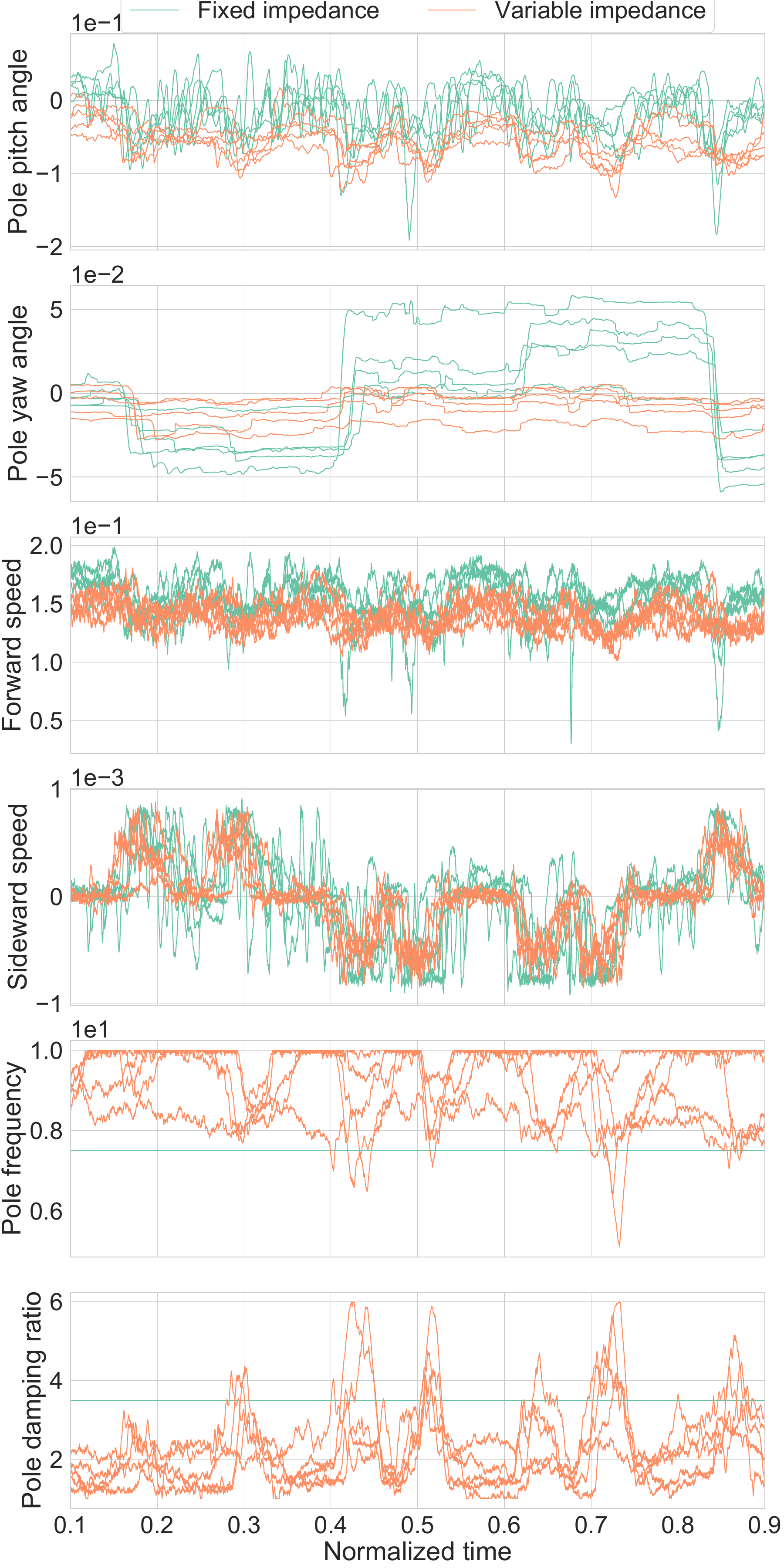}
    \caption{Experimental results.
    The fixed impedance controller had the worse operability, especially confirmed in the error of pole yaw angle.
    The variable impedance controller with the proposed method succeeded in stably operating the robot through five trials.
    }
    \label{fig:exp_result}
\end{figure}

Furthermore, variable impedance control is demonstrated using the proposed method, Accel MPC.
Seven parameters in the force-driven mobile robot are optimized to improve operability as six impedance parameters and one distribution ratio of the applied force between two impedance systems (see Appendix~\ref{app:task}).
As a basis, an impedance control that fixes the parameters on nominal values is also tested.
Since the improvement in optimization performance by Accel MPC has been statistically verified through the aforementioned simulations, we now turn our attention to real-time control on real robot.
The trajectories of the remarkable observations and parameters for five trials of operating the same course with each controller are summarized in Fig.~\ref{fig:exp_result}, which were obtained using six iterations in most steps.
Note that since there are unsuitable data without a user touching the robot for starting and stopping the trial from keyboard commands.
The trials are therefore evaluated using 10--90~\% of the time from the beginning to the end of the trial, which can be considered steady-state behavior with the user touching.
In addition, the attached video shows examples of the respective trials.

It can be seen that Accel MPC smoothed out the observations (top four of Fig.~\ref{fig:exp_result}).
Specifically, the fixed impedance control caused a problem in which the yaw axis of the pole did not return to its resting position after it moved.
This residual laterally disturbed even moving straight ahead.
In contrast, with the variable impedance control, the yaw axis of the pole moved only slightly near the origin.
This enables a stable straight operation with a nearly constant forward speed and almost zero sideward speed.
Nevertheless, the turning motion was not inconvenient and almost the same turning trajectories were drawn in all five trials.
These excellent operational performances are the outcome of optimization by Accel MPC, which can be seen in the bottom two of Fig.~\ref{fig:exp_result}.
Namely, during the straight operation, the posture of the pole was attempted to be fixed by increasing its natural frequency and decreasing its damping ratio.
At the turning motions, Accel MPC responded immediately by increasing the damping ratio and slightly suppressing the frequency.

However, not all trials converged with the same parameters.
This implies that the proposed method converged to one of the locally optimal solutions obtained earlier.
To make the output from the proposed method smooth for practicality, adding a smoothing function as a cost or a limit on the amount of variation in the output (although the latter was employed in this experiment) is desired.
Note that the warm start would also be effective for a smooth output, but it has a risk of overfitting.
Although such a restriction may reduce responsiveness, the high real-time control performance of the proposed method allows for a high optimization period (50~Hz in this experiment), and the original responsiveness is high enough that a slight reduction in responsiveness is not a problem.

\section{Conclusion}

This study proposed a methodology to improve the real-time performance of the versatile sampling-based MPC for more practical robot control.
Specifically, by replacing the minimization of the conventional FKL divergence with that of RKL divergence, a well-known mode-seeking property that can find a local optimal solution early was inherited theoretically.
The derived optimization problem, which was previously a weighted maximum likelihood estimation with weights limited to non-negative, evolved into a weighted maximum likelihood estimation with positive and negative weights.
We revealed that this can be solved using a (dynamic) MD algorithm.
While negative weights should eliminate unnecessary candidates, direct implementation may cause interference with positive and negative updates.
Therefore, a novel implementation that avoids interference was developed using pseudo-rejection sampling.
In addition, the latest NAG was modified for this problem to accelerate the convergence of MD.
Its adaptive step size was additionally designed to reduce the sensitivity to gradient noise.
The method finally obtained, the so-called Accel MPC, statistically outperformed the conventional method in Brax simulations.
It also exemplified the real-world applicability through optimizing the operability in a variable impedance control of a force-driven mobile robot.

While the proposal in this paper has made sampling-based MPC certainly more practical, several future works have been found through verification.
\begin{itemize}
    \item One is the ad-hoc design of $\bar{\pi}^-(U_t; \theta)$ in eq.~\eqref{eq:policy_not_bad}.
    A more theoretical design of the complementary distribution or formal implementation of rejection sampling that can guarantee constant computational cost is desirable.
    \item The output from MPC is one of the locally optimal solutions, raising a risk of large output variation at each time step.
    The restriction of output variation will be theoretically integrated with the optimization problem as in C/GMRES~\cite{ohtsuka2004continuation}.
    \item Tuning the hyperparameters, in particular the step size $\alpha$ (as it is cumulative in Accel MPC) and the horizon length $H$, is also a challenge, although most of them can be determined according to computational resources and implementation.
    We would like to develop a method that is easier or more robust to adjustment, referring to the latest optimizers developed for deep learning in recent years~\cite{zhang2019lookahead,ilboudo2023adaterm}.
    \item For simplicity, this paper assumed a diagonal normal distribution for the policy, but further improvement in efficiency can be expected by combining the proposed method with pre-learning of a distribution model with richer expressiveness~\cite{kusumoto2019informed,sacks2023learning}.
    However, in such a case, an analytical solution for KL divergence may not be available (unlike eq.~\eqref{eq:kl_normal}), so alternative constraints must be incorporated into MD.
    \item The hard-constrained problem is difficult to be directly solved using the sampling-based MPC, although the hard constraints can be converted into the large costs in practice.
    The most naive idea is to pool the candidates that satisfy the constraints and select the optimal one from them, but there is no guarantee of finding a safe candidate within the time limit.
\end{itemize}

\bmsection*{Author contributions}

T.K.: Conceptualization, Funding acquisition, Investigation, Methodology, Project administration, Software, Supervision, Validation, Visualization, Writing.
\\
K.F.: Investigation, Software, Validation, Visualization, Writing.

\bmsection*{Acknowledgments}

This work was supported by JST, PRESTO Grant Number JPMJPR20C3, Japan; and JST, CRONOS Grant Number JPMJCS24K6, Japan.

\bmsection*{Financial disclosure}

None reported.

\bmsection*{Conflict of interest}

The authors declare no potential conflict of interests.

\bibliography{biblio}

\bmsection*{Supporting information}

The video is attached to show the behaviors in simulations and real-robot demonstrations.

\appendix

\bmsection{Ablation tests}
\label{app:ablation}

\bmsubsection{Design of $\bar{\pi}^-(U_t; \theta)$}

First, in Reject MPC, the definition of $\bar{\pi}^-(U_t; \theta)$ in eq.~\eqref{eq:policy_not_bad} is verified.
For this purpose, the definition is once extended using $\kappa \geq 0$ as follows:
\begin{align}
    \bar{\pi}^-(U_t; \theta) &\propto \frac{1}{\pi^-(U_t; \theta^-) + \kappa \pi^+(\mu^-; \theta^+)}
    \label{eq:policy_not_bad_ablation}
\end{align}
In other words, if $\kappa=0$, it is purely for the complementary distribution of $\pi^-(U_t; \theta^-)$; and if $\kappa \to \infty$, $\pi^-(U_t; \theta^-)$ is no longer affected, so Reject MPC would revert to Forward MPC.

With this extended definition, the simulations in Section~\ref{subsec:sim1} with three patterns for $\kappa=0, 1, 10^5$ are shown in Table~\ref{tab:ablation}.
It is remarkable in $\kappa=0$ that \textit{reacher} failed because $\pi^-(U_t; \theta^-)$ and $\pi^+(U_t; \theta^+)$ were overlapped, rejecting the candidates that should be evaluated as the good ones.
In contrast, for $\kappa=10^5$, which is closer to Forward MPC, the performance of \textit{swimmer} was similarly low to Forward MPC.

From the above, it can be seen that adding $\pi^+(\mu^-; \theta^+)$ to the denominator moderately contributes to the generalization of Reject MPC.
As for the criterion for judging whether or not to reject the candidates near the center of the two overlapped distributions, the direction of inequality for them is a guideline: i.e. if $\pi^+(\mu^-; \theta^+)$ is smaller, it should be rejected; and if larger, it should be accepted.
Therefore, $\kappa=1$ is a natural choice, as defined in eq.~\eqref{eq:policy_not_bad} implicitly without clarifying the existence of $\kappa$.

\begin{table}[tb]
    \caption{Ablation test for pseudo-rejection sampling}
    \label{tab:ablation}
    \centering
    \begin{tabular}{cccc}
        \hline\hline
        $\kappa$ in eq.~\eqref{eq:policy_not_bad_ablation} & \multicolumn{3}{c}{Task} \\
        & \textit{acrobot} & \textit{reacher} & \textit{swimmer}
        \\
        \hline
        $\kappa=0$ & 891 $\pm$ 25 & \textbf{-12.08} $\pm$ 3.66 & 23.5 $\pm$ 3.9
        \\
        $\kappa=1$ & 904 $\pm$ 19 & -2.34 $\pm$ 0.72 & 23.5 $\pm$ 4.0
        \\
        $\kappa=10^5$ & 914 $\pm$ 20 & -2.16 $\pm$ 0.68 & \textbf{22.7} $\pm$ 4.2
        \\
        \hline\hline
    \end{tabular}
\end{table}

\bmsubsection{Choice of sample scales calculated}

Next, the selection of the sample scales to be used for the calculation of $s_i$ defined in eq.~\eqref{eq:noise_strength} is investigated.
Indeed, using the scales of the gradient itself is most direct, but from the viewpoint of faster implementation, it was better to derive the updated results without obtaining the gradient, so the gradient was not calculated explicitly in the implementation.
In addition, the gradient scales need to be given for each component of $\theta^{+,-}$, wasting the computation and memory costs.

As a compromise, the scales for the input $U$ and output $J$ of $J(x, U)$, which are involved in the gradient calculation, are considered.
On one hand, for $U$, its scales are in $\mathbb{R}_+^{|\mathcal{U}| \times H}$, namely the scales should be computed for each action dimension and horizon, while their maximum $\sigma^{\mathrm{max}}$ would be suggested as $\sigma_1 (= 1)$.
On the other hand, for $J$, it is sufficient to compute the common $s_i$ for all components of $\theta^{+,-}$ due to $\sigma \in \mathbb{R}_+$, while $\sigma^{\mathrm{max}}$ is task-dependent.
Therefore, $\sigma^{\mathrm{max}}$ must be numerically estimated and stored in each optimization.
If $\sigma^{\mathrm{max}} = \max_{k=1, \ldots, i} \sigma_k^{\mathrm{std}}$, outliers in $J$ will have a negative impact, so $\sigma^{\mathrm{max}} = \max_{k=1, \ldots, i} \sigma_k^{\mathrm{mad}}$ is employed.

\begin{figure}[tb]
    \centering
    \includegraphics[keepaspectratio=true,width=0.96\linewidth]{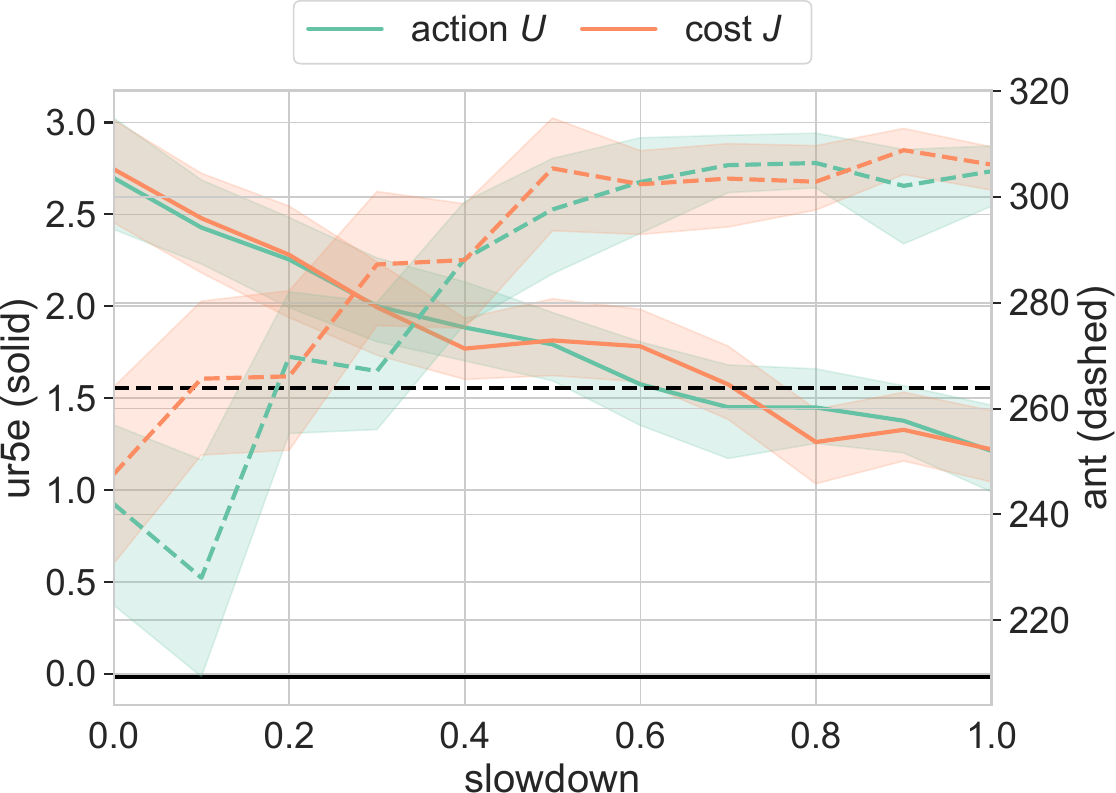}
    \caption{Ablation test for the adaptive step size.
    There is no significant difference between the variables $U$ and $J$ used in the scale calculation.
    The fine tuning of $\gamma$ depends on the task to be solved.
    }
    \label{fig:sim_ablation_slowdown}
\end{figure}

The adaptive step sizes with $\gamma=0, 0.1, \ldots, 1, 100$ using these two scales were tested in the simulations in Section~\ref{subsec:sim2} without the warm start for simplicity and are summarized in Fig.~\ref{fig:sim_ablation_slowdown}.
Here, $\gamma=100$ is shown as black horizontal lines as reference, since it is set so that no acceleration occurs.
There was no significant difference in the performances regardless of which scale is used.
In other words, the appropriate $\gamma$ differs depending on the task.
There seems to be a division between simple tasks that do not require slowdown (e.g. \textit{ur5e}) and tasks that require a certain degree of careful updating (e.g. \textit{ant}).
However, the performance was improved compared with $\gamma=100$, confirming that the acceleration by the modified AGD+ is effective.
Although a more optimal design for this point is an open issue, this study employs the scales of $J$, which has achieved a more stable performance.

\bmsection{Toy problem}
\label{app:toy}

Here, a simple toy problem is conducted for showing the different convergence properties with the conventional method and the proposed method, as suggested in Fig.~\ref{fig:diff_fkl_rkl}.
Specifically, it contains a two-dimensional point agent, which has an omnidirectional velocity controller with a deadzone.
That is, the action is a two-dimensional velocity command $u \in [-0.5, 0.5]^2$ and the agent aims to reach a goal located at $(1, 1)$ from a starting point near $(0, 0)$.
The deadzone is a discrete process in which each dimension of $u$ is resetted to zero if its magnitude is less than $0.1$.
The reward is the sum of i) a term inversely proportional to the distance between the goal and agent; and ii) a term inversely proportional to the magnitude of $u$.
A circular obstacle with a radius of 0.5~m is placed on $(0.5, 0.5)$.

\begin{figure}[tb]
    \centering
    \includegraphics[keepaspectratio=true,width=0.96\linewidth]{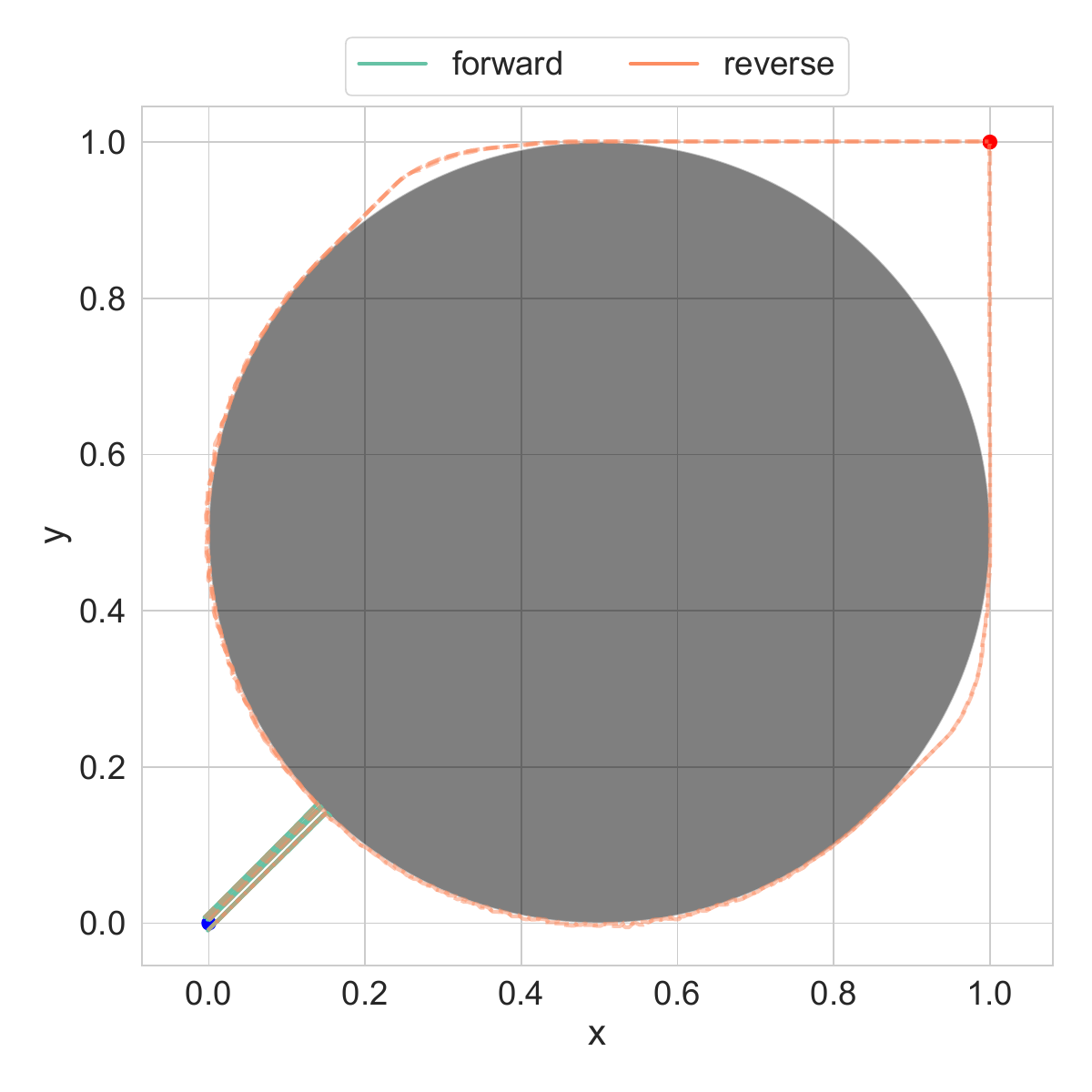}
    \caption{Trajectories in toy problem.
    The conventional method with forward KL divergence minimization collides with the obstacle.
    The proposed method with reverse KL divergence minimization selects either of the fork.
    }
    \label{fig:toy_trajectory}
\end{figure}

\begin{figure}[tb]
    \centering
    \includegraphics[keepaspectratio=true,width=0.96\linewidth]{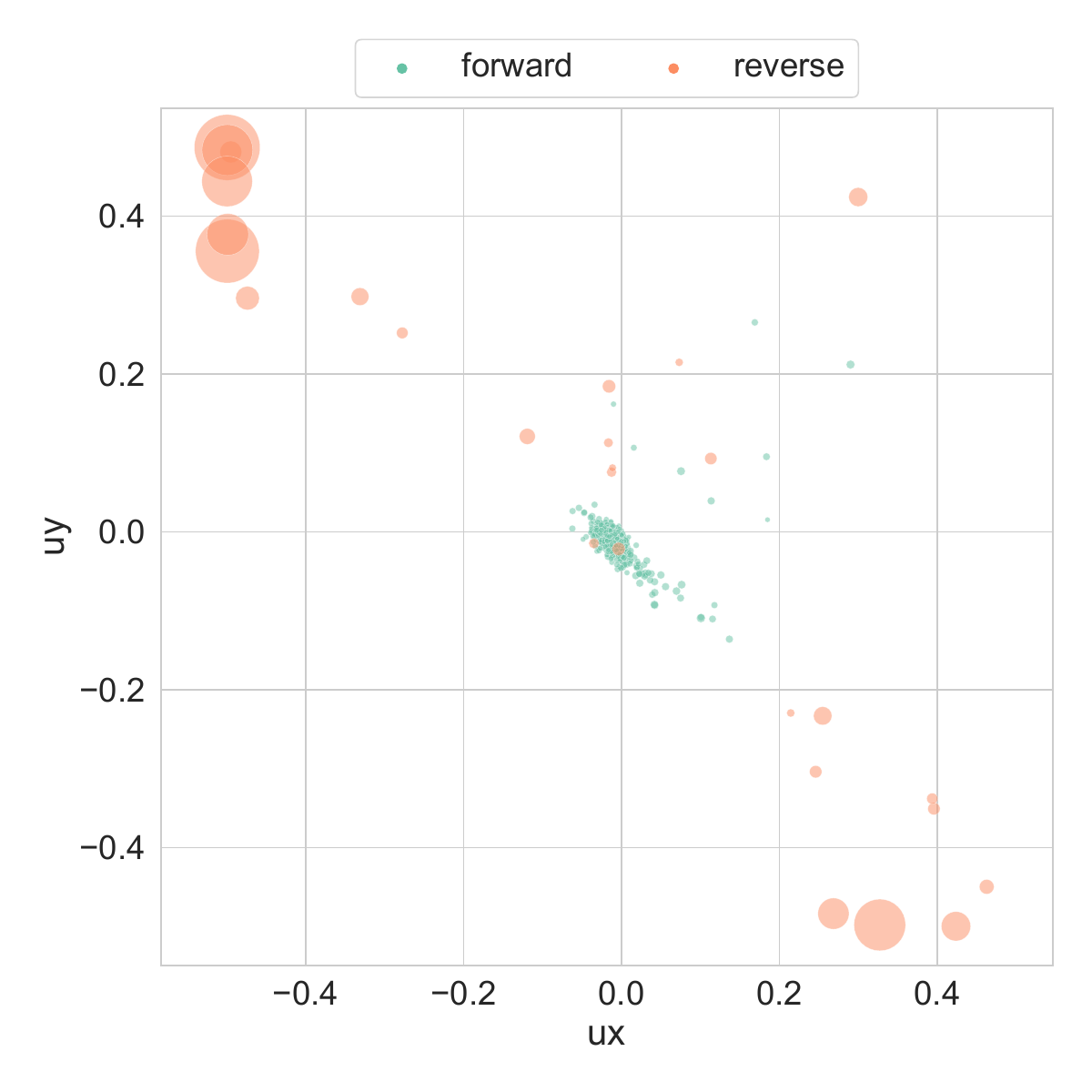}
    \caption{Policies in the vicinity of junction point.
    The forward method made the policy converge to the center of the optimal bimodal policy.
    The reverse method prefentially obtained either of the modes of the optimal policy.
    }
    \label{fig:toy_policy}
\end{figure}

To confirm the behavior of a well-converged MPC policy for this toy problem, optimization was performed over 25 iterations without any time limit.
The trajectories of five trials of the conventional and proposed methods (\textit{forward} and \textit{reverse}, respectively) are shown in Fig.~\ref{fig:toy_trajectory}, and policies near a junction point (i.e. $(0.5 - 0.5/\sqrt(2), 0.5 - 0.5/\sqrt(2))$) are illustrated in Fig.~\ref{fig:toy_policy}.
As can be seen from the results, at the junction point, \textit{forward} was stuck in place, failing to initiate either a clockwise or counter-clockwise path.
This is because the action candidates for the respective paths canceled each other out and the policy converged to the action of stopping in place, which could obtain the term for minimizing $u$.
In contrast, \textit{reverse} successfully avoided the obstacle and passed through one of the clockwise and counter-clockwise paths without staying at the junction point.
The policy at the junction point was biased in one direction or another, confirming the mode-seeking property.

\bmsection{Details of tasks}
\label{app:task}

\bmsubsection{Simulation tasks}

As mentioned above, Brax~\cite{freeman2021brax} was used for the simulator.
The benchmark tasks provided in the library, \textit{acrobot}, \textit{reacher}, \textit{swimmer}, \textit{ur5e}, and \textit{ant}, were used for verification in this study.
The first three are relatively simple tasks and targeted in Section~\ref{subsec:sim1}, whereas the remaining two are more complex robot tasks and targeted in Section~\ref{subsec:sim2}.
The PC running these simulations was equipped with a CPU, Intel Core i9-10980XE, and a GPU, NVIDIA GeForce RTX 3080.

In these tasks, the number of internal action repetitions was adjusted so that the control cycle became 10~Hz, and one trial was completed in a maximum of 12~seconds.
Each task has an original reward (i.e. negative cost) function, and the first three tasks used the original as-is, whereas the remaining two tasks were slightly modified.
Specifically, in \textit{ur5e}, the judgment for reaching the target was changed to a radius of 0.05~m.
In \textit{ant}, the cost of the magnitude of the action was eliminated, and the cost of avoiding contact with the ground was reversed as a bonus to walk with sliding feet.

\bmsubsection{Force-driven mobile robot}

For the demonstrations, a force-driven mobile robot~\cite{kobayashi2022light} with a two-axis pole on top of a two-wheeled differentially-driven mobile base was used.
This pole is equipped with two actuators developed by HEBI Robotics, which can measure the applied torque.
Each pole and base has its own impedance controller with a virtual mass, spring, and damper.
In addition, these two are coupled with each other: for example, the rest length of the base depends on the pole position and the torque applied to the pole is distributed to the base at the specified distribution ratio.
This system is controlled with 50~Hz only using CPU, Intel Core i3-8109U.

MPC optimizes seven dimensions of the impedance parameters (more specifically, the virtual mass, natural frequency, and damping ratio for the two impedance systems) and the total torque distribution ratio.
In other words, the action space of MPC is defined as the amount of change in these seven dimensions.
The corresponding state has 15 dimensions: the pole angle, angular velocity, torques, base planar velocity, and current parameters.
The cost function has three types of terms: the stabilization term to make the parameters nominal, the motion smoothing term to match the pole movement with the base movement, and the constraint term for the pole angle and base speed.
However, in the demonstrations, the last constraint term was hardly activated; therefore, the first two terms were in fact optimized.

\begin{figure}[tb]
    \centering
    \includegraphics[keepaspectratio=true,width=0.96\linewidth]{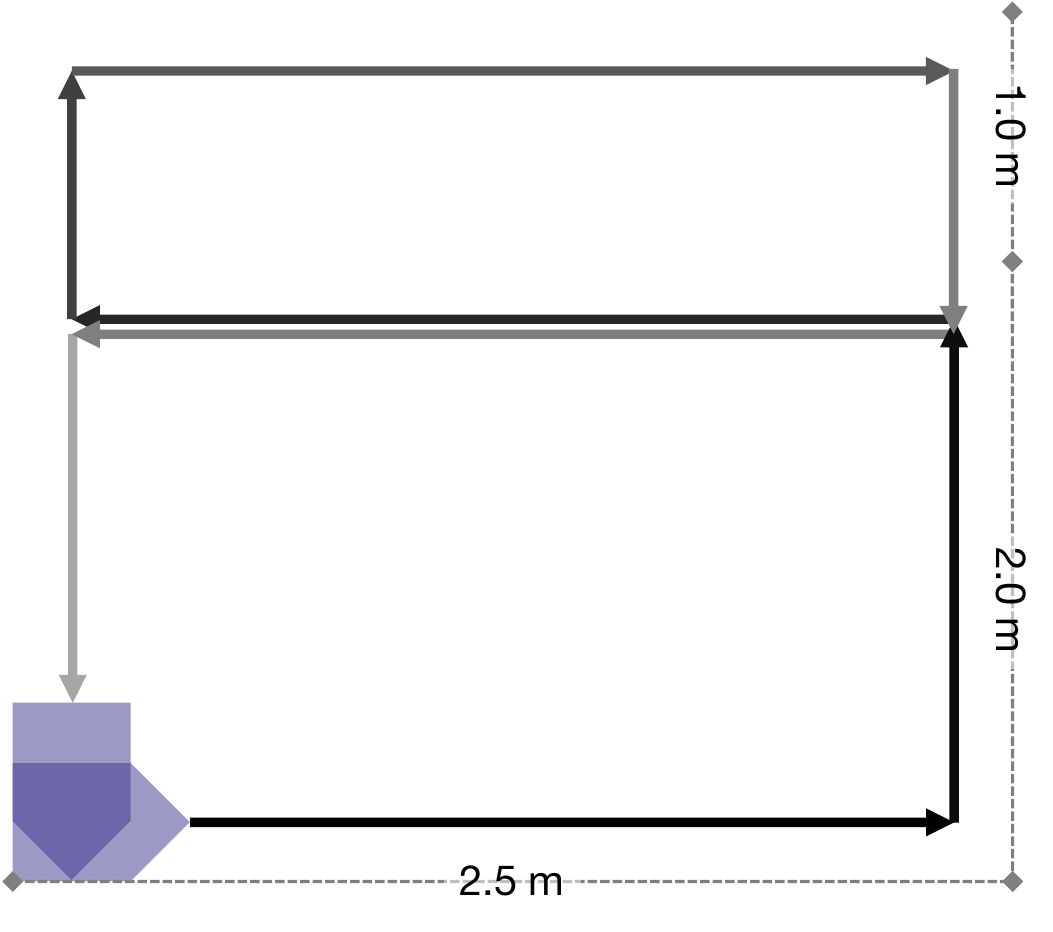}
    \caption{Experimental course.
    The force-driven robot is operated by following the arrows from the bottom left to draw a figure eight.
    }
    \label{fig:exp_course}
\end{figure}

The robot was operated to move around a course as illustrated in Fig.~\ref{fig:exp_course}.
This course includes all basic motions, i.e. going straight, turning left, and right.
The user of this robot aimed to maintain a constant forward speed when going straight and turning at right angles as much as possible.

\bmsection{Warm start}
\label{app:warm}

In general, the warm start reuses $\theta^{\ast}$ of the optimization results obtained at $t-1$ to $\theta_1$ at $t$ time step.
In this study, the warm ratio $\eta \in [0, 1]$ is introduced and initialized as follows, taking into account the lack of sufficient optimization and the influence of modeling errors.
\begin{align}
    \theta_1[0:H-1] \gets (1 - \eta) \theta_1[0:H-1] + \eta \theta^{\ast}[1:H]
\end{align}
where $x[y:z]$ extracts the $y$-to-$z$ components of $x$.

In addition to this standard warm start, Accel MPC also wants to reuse the information accumulated in the previous optimization for $a_i$ and $A_i$.
Therefore, $a_1$ is first defined by interpolating its initial value $\alpha$ and the previous final value $a_{\mathrm{prv}}$ as follows:
\begin{align}
    a_1 = (1 - \eta) \alpha + \eta a_{\mathrm{prv}}
\end{align}
This can be interpreted as having done up to the $i = a_1 / \alpha$-th iteration under the original definition, $a_i = \alpha i$.
Based on this interpretation, $A_1$ is warmly obtained by substituting it into $A_i = \sum_{k=1}^i a_k = \alpha i (i + 1) / 2$.
\begin{align}
    A_1 = \frac{a_1}{2} \left( \frac{a_1}{\alpha} + 1 \right)
\end{align}


\end{document}